%% file: main.tex
\documentclass[10pt,twocolumn,letterpaper]{article}
\PassOptionsToPackage{table}{xcolor}
\usepackage{cvpr}              
\input{preamble}

\definecolor{cvprblue}{rgb}{0.21,0.49,0.74}
\usepackage[pagebackref,breaklinks,colorlinks,allcolors=cvprblue]{hyperref}

\def\paperID{1211} 
\def\confName{CVPR}
\def\confYear{2025}

\title{Vision-Language Model IP Protection via Prompt-based Learning}

\author{Lianyu Wang$^1$\footnotemark[1], \; Meng Wang$^2$\footnotemark[1], \; Huazhu Fu$^3$\footnotemark[2], \; Daoqiang Zhang$^1$\footnotemark[2] \\
\small{$^1$The Key Laboratory of Brain-Machine Intelligence Technology, Ministry of Education}\\ 
\small{$^2$Centre for Innovation and Precision Eye Health, Yong Loo Lin School of Medicine, National University of Singapore}\\ 
\small{$^3$Institute of High Performance Computing, Agency for Science, Technology and Research}
}

\begin{document}
\maketitle
\input{sec/0_abstract}  
\renewcommand{\thefootnote}{\fnsymbol{footnote}} 
\footnotetext[1]{L.~Wang and M.~Wang contributed equally to this work.} 
\footnotetext[2]{Corresponding author: H.~Fu (hzfu@ieee.org) and D.~Zhang (dqzhang@nuaa.edu.cn).} 
\input{sec/1_intro}

\input{sec/2_related}

\input{sec/3_method}
\input{sec/4_exp}
\input{sec/5_conclusion}
\section*{Acknowledgments}
This work is supported by the National Natural Science Foundation of China (Nos. 62136004, 62276130), the Key Research and Development Plan of Jiangsu Province (No. BE2022842), and H. Fu’s A*STAR Central Research Fund.
{
    \small
    \bibliographystyle{ieeenat_fullname}
    \bibliography{ref}
}

\input{sec/X_suppl}
\end{document}

%% file: preamble.tex
\usepackage{graphicx}
\usepackage{amsmath}
\usepackage{amssymb}
\usepackage{booktabs}
\usepackage{algorithm}
\usepackage{algorithmic}
\usepackage{makecell}
\usepackage{multirow}

%% file: sec/0_abstract.tex
\begin{abstract}
Vision-language models (VLMs) like CLIP (Contrastive Language-Image Pre-Training) have seen remarkable success in visual recognition, highlighting the increasing need to safeguard the intellectual property (IP) of well-trained models. Effective IP protection extends beyond ensuring authorized usage; it also necessitates restricting model deployment to authorized data domains, particularly when the model is fine-tuned for specific target domains. However, current IP protection methods often rely solely on the visual backbone, which may lack sufficient semantic richness.
To bridge this gap, we introduce IP-CLIP, a lightweight IP protection strategy tailored to CLIP, employing a prompt-based learning approach. 
By leveraging the frozen visual backbone of CLIP, we extract both image style and content information, incorporating them into the learning of IP prompt.
This strategy acts as a robust barrier, effectively preventing the unauthorized transfer of features from authorized domains to unauthorized ones.
Additionally, we propose a style-enhancement branch that constructs feature banks for both authorized and unauthorized domains. This branch integrates self-enhanced and cross-domain features, further strengthening IP-CLIP’s capability to block features from unauthorized domains.
Finally, we present new three metrics designed to better balance the performance degradation of authorized and unauthorized domains. 
Comprehensive experiments in various scenarios demonstrate its promising potential for application in IP protection tasks for VLMs.
\end{abstract}

%% file: sec/1_intro.tex
\section{Introduction}
\label{sec:intro}

\begin{figure}[!t]
  \centering
   \includegraphics[width=1\linewidth,trim=180 220 180 210,clip]{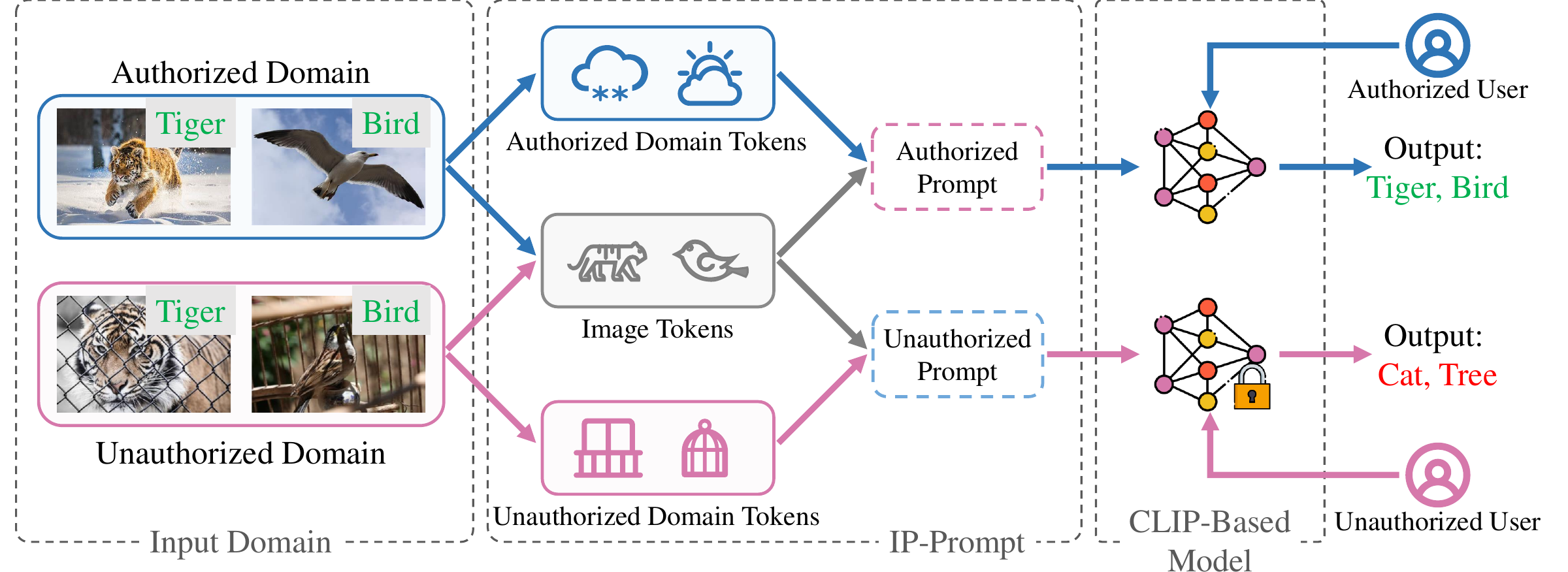}
   \caption{Illustration of model IP protection with IP-CLIP. Domain and image tokens form the IP-Prompt, which a CLIP-based model audits to verify data origin. This prevents unauthorized transfers and degrades performance in unauthorized domains. Notably, IP-Prompt is a lightweight, plug-and-play module for CLIP-based models.}
   \label{figure1}
\end{figure}

Driven by the availability of large-scale data and powerful computing hardware, vision-language models (VLMs) like CLIP have recently achieved remarkable generalization across a wide range of downstream tasks~\cite{clip,CoOp,CoCoOp}, leading to a surge in their commercial significance. However, developing a well-trained VLM is a resource-intensive endeavor, requiring substantial investments in time, manpower, and resources. This includes the design of specialized architectures~\cite{ResNet,ViT}, access to vast amounts of high-quality data~\cite{SAM,Imagenet,wang2023uncertainty}, and the use of expensive computational resources~\cite{NAS}. As a result, protecting these models' intellectual property (IP) has garnered significant attention~\cite{IP,NTL,CUTI,CUPI}.

Previous research on IP protection has primarily concentrated on two aspects: ownership verification (i.e., verifying who owns the model)~\cite{peng,bai,ren} and usage authorization (i.e., authorizing who has the right to deploy the model)~\cite{usage1,usage2}. Some of these approaches incorporate deep watermarks, embedding unique identifiers such as inputs, parameters, gradients, architectures, or even outputs. Others extract distinctive model characteristics, acting as “fingerprints”~\cite{fingerprints} for deep models. 
While these techniques provide a degree of protection, they can be easily bypassed through fine-tuning or retraining. Moreover, authorized users are often unrestricted in how they apply the model, allowing them to effortlessly transfer high-performance models to similar tasks, which can lead to implicit IP infringement. 
This problem stems from the fact that VLM’s trained visual backbones often generalize across domains, which can breed model stealing, leading to illegal misuse and implicit intellectual property infringement.
An intuitive solution is to refine the model’s generalization boundary to focus on domain-specific features and restrict their use to authorized domains. NTL~\cite{NTL} achieves this by amplifying the maximum mean discrepancy (MMD) between authorized and unauthorized domains, thus narrowing the model’s generalization scope. In contrast, CUTI-domain~\cite{CUTI} introduces an intermediate domain that combines features from both domains, preventing unauthorized transfers.
Although existing deep model IP protection methods can provide commendable performance in specific scenarios, they face two fundamental challenges. Firstly, they require training models from scratch or extensive fine-tuning, which is particularly demanding for VLMs due to their resource-intensive nature. To address this, some prompt tuning methods techniques, such as CoOp~\cite{CoOp} and MaPLe~\cite{MaPLe} have shown superior performance on some specific downstream tasks. CoOp uses soft prompts to learn text prompts, while MaPLe introduces visual language prompts to enhance synergy.
Secondly, some methods~\cite{CUTI,NTL} attempt to constrain model performance by generating supplementary data. However, these methods often introduce additional training steps, and the generated data typically lack adequate constraints and control, complicating practical use.

To tackle these challenges, we introduce IP-CLIP, a novel approach for IP protection in CLIP-based models. IP-CLIP utilizes a lightweight prompt-tuning technique called IP-Prompt (illustrated in~\cref{figure1}) to distinguish between authorized and unauthorized prompts without requiring full fine-tuning of all pre-trained parameters. Our approach involves learning new prompts consisting of two types of tokens: i) Authorized/unauthorized domain token: this token captures the multi-scale style information of authorized/unauthorized domains from the CLIP visual encoder. ii) Image token: to effectively learn the visual distribution in the semantic space and obtain cue distributions for each class, we utilize multi-scale visual feature responses from various layers of the CLIP visual encoder. The downstream CLIP-based model integrates these two tokens into its decision-making process, allowing it to simultaneously identify both the Authorization and category of the input image. This enables accurate predictions for images from the authorized domain while deliberately producing incorrect results for samples from unauthorized domains. Notably, IP-Prompt functions as a lightweight, plug-and-play module that can be positioned at the front end of various CLIP-based models to provide IP protection.
Additionally, we introduce a style enhancement branch with feature banks for both authorized and unauthorized domains. This branch integrates self-enhanced and cross-domain features into the model, improving its ability to recognize authorized features while excluding unauthorized ones.
Finally, we design three new metrics tailored to the IP protection scenario to balance performance between authorized and unauthorized domains.
The main contributions of this paper are summarized as follows:

\begin{itemize}
\item{We propose the \textbf{IP-CLIP} framework, an innovative approach for IP protection of VLMs, with only minimal parameter updates. This framework is designed to prevent the unauthorized transfer of well-trained, large-scale VLMs from authorized to unauthorized domains.}
\item{We design a lightweight, plug-and-play \textbf{IP-Prompt} that can be integrated into various CLIP-based models for effective IP protection of VLMs.}
\item{Our approach includes a \textbf{style enhancement branch} that generates diverse visual features and integrates self-enhanced and cross-domain features into the model. This enables the protected model to better identify authorized features and exclude unauthorized ones.}
\item{We introduce three \textbf{new metrics} for a comprehensive evaluation of IP protection capabilities, addressing gaps in current methods. Extensive experiments demonstrate the effectiveness of IP-CLIP on various datasets and scenarios, providing strong evidence that our method offers a robust solution for model IP protection.\footnote[1]{https://github.com/LyWang12/IP-CLIP}}
\end{itemize}

%% file: sec/2_related.tex
\section{Related Work}
\label{sec:related}

\subsection{Visual Language Models and Prompt Tuning}
Large-scale visual language models (VLMs) integrate visual and textual inputs for a more comprehensive understanding, achieving strong performance in various computer vision tasks~\cite{VLM1,VLM2,VLM3}. Models like CLIP~\cite{clip} and VisualBERT~\cite{visualbert} rely on pre-trained language models (e.g., BERT~\cite{bert}, GPT~\cite{gpt}) for text encoding, while visual inputs are processed via convnets or visual transformers. As these models scale up, their computational demands increase, making updates costly. To address this, parameter-efficient tuning methods are essential.

Prompt tuning is one such approach, which focuses on learning a small set of parameters while keeping the larger model frozen~\cite{VPT}. CoOp~\cite{CoOp} introduced the use of soft prompts in VLMs, demonstrating that carefully crafted text prompts can enhance image recognition performance. By incorporating lightweight neural networks to dynamically generate prompts for individual images, CoCoOp~\cite{CoCoOp} addresses the issue of prompt overfitting. VPT~\cite{VPT} achieved strong results by using a small number of visual prompts, and MaPLe~\cite{MaPLe} further combined textual and visual prompts within CLIP to improve the alignment between text and image representations. Although these parameter fine-tuning methods have demonstrated effectiveness, they offer insufficient security. Lacking robust IP protection, the critical issue of safeguarding IP in large-scale models has garnered growing attention and scrutiny.

\subsection{Intellectual Property (IP) Protection}
A comprehensive IP protection strategy should address both ownership verification and applicability authorization. Ownership verification identifies the rightful owner of the model, typically using watermarks or fingerprinting. Peng~\etal~\cite{peng} introduced a general adversarial perturbation fingerprinting method, which uses contrastive learning to match fingerprints with similarity scores.  Bai~\etal~\cite{bai} proposed BadCLIP, which impacts image and text encoders using trigger-aware prompts, while. Ren~\etal~\cite{ren} adopted a poison-only backdoor approach for embedding watermarks and used hypothesis testing for remote verification. However, these methods have been proven vulnerable to certain removal and covering techniques.

Applicability authorization focuses on restricting the model's generalizability to specific domain. Wang~\etal~\cite{NTL} introduced non-transfer learning (NTL), which uses an estimator with a feature kernel to highlight domain-specific differences. Zeng~\etal~\cite{zeng} extended NTL to natural language processing and auxiliary domain classifiers for better domain separation. Hong~\etal~\cite{HNTL} further proposed H-NTL, leveraging a causal model to disentangle content and style as latent factors, thereby guiding the learning of non-transferable representations based on intrinsic causal relationships.
Wang~\etal~\cite{CUTI} proposed an innovative compact non-transferable isolation domain (CUTI-domain) to isolate authorized and unauthorized domains, limiting performance transfer. Existing IP protection methods can be effective but often require extensive training or fine-tuning, which is resource-intensive for VLMs. Additionally, methods relying on supplementary data often lack necessary constraints and controllability, complicating their practical use.

%% file: sec/3_method.tex
\section{Method}
\label{sec:method}

\subsection{Problem Definition}
IP protection aims to confine model performance to the authorized domain while reducing its recognition ability in the unauthorized domain. Formally, we define the IP protection task as follows~\cite{define}:

\textbf{Definition 1} (IP protection): \textit{Let $D_a={\left\{x_{ai}, y_{ai}\right\}}_{i=1}^{N_a}$ denote the dataset for the authorized domain, and $D_u={\left\{x_{ui}, y_{ui}\right\}}_{i=1}^{N_u}$ represent the dataset for the unauthorized domain, where $N_a$ and $N_u$ are the number of samples in the authorized and unauthorized domains, respectively. Data $X_a$ and $X_u$ from these domains are drawn from different distributions but share the same label space $Y$. In the authorized domain, the model aims to map data to labels:
\begin{equation}
\label{d1}
F(X_a) \rightarrow Y.
\end{equation} 
The challenge of the IP protection task is to achieve non-transferability to the unauthorized domain while minimally affecting performance in the authorized domain:
\begin{equation}
\label{d2}
F(X_u) \perp Y~and~F(X_a) \perp F(X_u), 
\end{equation} 
where $\perp$ denotes statistical independence.} Current IP protection methods usually rely solely on visual backbones~\cite{CUTI,NTL,define}, which may lack sufficient semantic richness. To bridge this gap, we introduce IP-CLIP, a lightweight IP protection strategy tailored for vision-language models.

\begin{figure*}[t]
  \centering
   \includegraphics[width=0.9\linewidth,trim=135 105 130 105,clip]{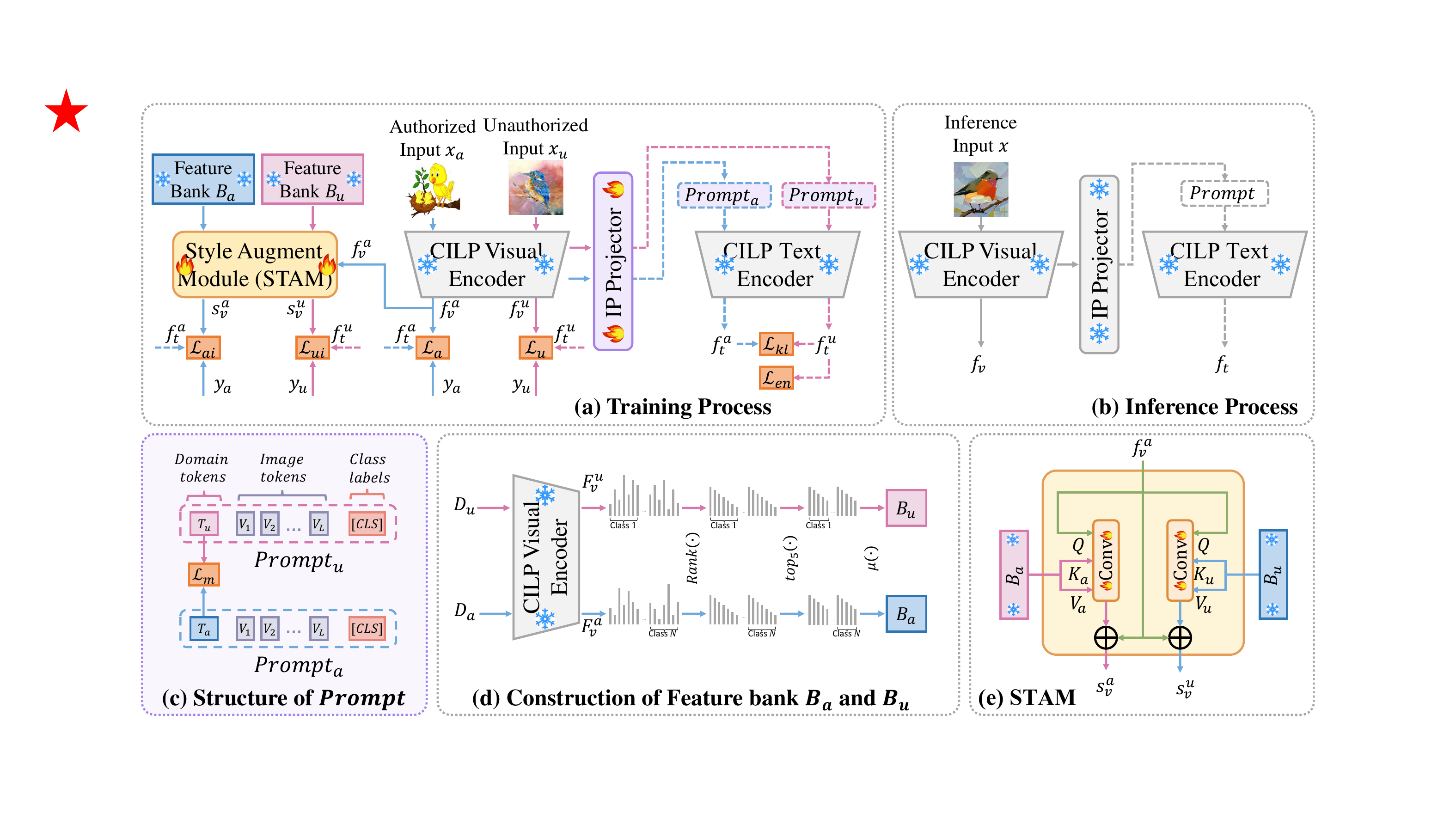}
   \caption{(a) The architecture of IP-CLIP is based on a frozen CLIP backbone, where snowflakes denote frozen layers and sparks represent trainable layers. During training, inputs from both the authorized domain $x_a$ and unauthorized domain $x_u$ are fed into the frozen CLIP visual encoder in parallel to generate feature vectors $f_v^a$ and $f_v^u$. The IP projector extracts domain tokens and image tokens from the visual encoder, which are then used to construct prompts as inputs to the text encoder. The style enhancement branch takes the frozen feature bank and $f_v^a$ as input, with $s_v$ representing the enhanced visual features. The prediction result is derived by calculating the similarity between the visual feature $s_v$/$f_v$ and the text feature $f_t$. $y$ and $\mathcal{L}$ represent the label and loss function, respectively. (b) The Inference process of IP-CLIP. (c) Structure of ${Prompt}_a$ and ${Prompt}_u$. (d) Construction of Feature bank $B_a$ and $B_u$, where $D$ and $F$ represent the input dataset and its corresponding visual feature set, respectively. During training, the feature banks remain frozen. (e) Structure of STAM.} 
   \label{figure2}
\end{figure*}

\subsection{Overview of IP-CLIP}
\cref{figure2} (a) illustrates the details of our proposed IP-CLIP framework. The primary objective is to constrain model performance to the authorized domain by learning both image and domain-specific tokens, thereby emphasizing the unique features of the authorized domain while preventing unauthorized generalization. To accomplish this, we feed both the authorized domain data $x_a$ and the unauthorized domain data $x_u$ into CLIP's frozen visual encoder in parallel, producing the output features $f_v^a$ and $f_v^u$, respectively. A learnable IP Projector is employed to capture multi-scale features from different layers of the visual encoder, generating authorized / unauthorized domain tokens $T_a$ / $T_u$ and image tokens $[V_1, V_2, \ldots, V_L]$, which are concatenated as input prompts for the frozen text encoder of CLIP, as described in~\cref{prompt}. The prediction result is obtained by calculating the similarity between text feature $f_t$ and visual feature $f_v$, and the label is denoted as $y$. The style enhancement branch (\cref{branch}), associated with the feature banks, further improves the robustness of the features in distinguishing between authorized and unauthorized domains. The frozen layers of our proposed IP-CLIP framework are labeled with snowflakes, while the few trainable layers are marked with sparks. 

\subsection{Our Proposed Prompt Learning}\label{prompt}
Instead of the static prompting technique, we aim to learn prompts directly from the visual domain to efficiently encode visual distributions. Our IP protection approach has two main objectives in prompt tuning: i) introduce domain-specific tokens for authorized / unauthorized domains, and ii) generate domain-independent image tokens for visual recognition tasks, as illustrated in~\cref{figure2} (c). Specifically, multi-scale features $[f_v^{(1)}, f_v^{(2)},\ldots, f_v^{(M)}]$ are extracted from the frozen visual encoder, where $f_v^{(m)}$ represents the response from the $m$-th layer of the encoder. To create domain-specific tokens for authorized / unauthorized domains, multi-scale style features (represented by first-order and second-order batch-wise feature statistics) are computed and combined, resulting in $[\mu^{(1)};\sigma^{(1)};\ldots;\mu^{(M)};\sigma^{(M)}]$, which are then processed by the IP Projector to produce domain-specific tokens $T$. Additionally, the multi-scale features $[f_v^{(1)}, f_v^{(2)},\ldots, f_v^{(M)}]$ are passed through IP Projector to generate $L$ image-specific tokens $[V_1,V_2,\ldots,V_L]$. Finially, the prompt for the authorized domain is denoted as:
\begin{equation}
\label{eq1}
{Prompt}_a = [T_a;V_1,V_2,\ldots,V_L;[CLS]],
\end{equation} 
while for the unauthorized domain, it is denoted as:
\begin{equation}
\label{eq2}
{Prompt}_u = [T_u;V_1,V_2,\ldots,V_L;[CLS]],
\end{equation} 
which are then input into the frozen text encoder to generate text features $f_t^a$ and $f_t^u$, respectively.

\subsection{Style-Enhancement Branch}\label{branch}
For the style enhancement branch, we construct feature banks for both the authorized and unauthorized domains and introduce a style augment module (STAM) to diversify the features.

\textbf{Constructing feature banks.}
Leveraging CLIP's zero-shot capabilities, we extract text and image features from $D_a$ and $D_u$, as in~\cref{figure2} (d). For the authorized domain, we compute a confidence score (i.e., the maximum probability) for each image based on CLIP’s predictions. Similarly, in the unauthorized domain, we calculate confidence scores and assign pseudo-labels based on the highest score. We then select the visual features with the highest confidence in each category from both domains to construct $N$-way $K$-shot feature banks, where $N$ is the number of categories and $K=5$ is the number of samples per category. Finally, the centroid features for each category are calculated to form the authorized domain feature bank ($B_a$) and the unauthorized domain feature bank ($B_u$), both expressed as $\mathbb{R}^{N \times C}$, where $C$ denotes the feature dimension. Note that the feature bank is built by iterating over the data only before training, after which it is frozen during the training process.

\textbf{STyle Augment Module (STAM).} 
STAM utilizes the frozen feature banks to guide images in acquiring self-enhanced and cross-domain features, as illustrated in~\cref{figure2} (e). First, the query $Q$ is calculated from the input feature $f_v^a$, while the key $K_a$ and value $V_a$ are derived from the authorized domain bank. Similarly, $K_u$ and $V_u$ are calculated from the unauthorized domain bank.
We derive enhanced $s_v^a$ and $s_v^u$ by utilizing a learnable attention layer combined with a residual connection. This mechanism enables the image feature to concentrate on the features from the authorized or unauthorized domain banks. This process can be formally expressed as:
\begin{equation}
\label{eq3}
s_v^a = \text{Conv}(\text{softmax}\left(\frac{QK_a^T}{\sqrt{d_k}}\right) V_a) + f_v^a,
\end{equation} 
\begin{equation}
\label{eq4}
s_v^u = \text{Conv}(\text{softmax}\left(\frac{QK_u^T}{\sqrt{d_k}}\right) V_u) + f_v^u.
\end{equation} 
Here, $\sqrt{d_k}$ denotes the scaling factor, while $T$ represents the transpose operation.

\subsection{Training Strategy}\label{training}
\textbf{Target-specified IP-CLIP.}
We begin by detailing the training process for our proposed IP-CLIP, assuming both the authorized and unauthorized domains are known. To allow the model to effectively differentiate between the authorized domain token $T_a$ and unauthorized domain token $T_u$, we use mean squared error (MSE) loss to maximize their separation, as described by: 
\begin{equation}
\label{eq5}
\mathcal{L}_m = \mathcal{L}_{MSE}(T_a, T_u).
\end{equation} 
Next, we utilize contrastive loss function $\mathcal{L}_a$ / $\mathcal{L}_v$ to optimize the image-text mapping between image feature $f_v^a$ / $f_v^u$ and the text feature $f_t^a$ / $f_t^u$, as shown in:
\begin{equation}
\label{eq6}
\mathcal{L}_a = \frac{\text{exp}(\langle f_v^a, f_t^a(y_a)\rangle/\tau)}{\sum_{k=1}^{K} \text{exp}(\langle f_v^a, f_t^a(k)\rangle/\tau)},
\end{equation} 
where $\tau$ denotes temperature parameter, $K$ denotes the number of classes and $\langle \cdot, \cdot \rangle$ denotes the cosine similarity.

Similarly, the enhanced feature $s_v^a$ / $s_v^u$ is aligned with the text representation $f_t^a$ / $f_t^u$ by $\mathcal{L}_{ai}$ / $\mathcal{L}_{ui}$,  which can be expressed as:
\begin{equation}
\label{eq7}
\mathcal{L}_{ai} = \frac{\text{exp}(\langle s_v^a, f_t^a(y_a)\rangle/\tau)}{\sum_{k=1}^{K} \text{exp}(\langle s_v^a, f_t^a(k)\rangle/\tau)}.
\end{equation} 

For text representations, we use Kullback-Leibler (KL) divergence loss to further separate the distances between the authorized and unauthorized domains:
\begin{equation}
\label{eq8}
\mathcal{L}_{kl} = KL(f_t^a, f_t^u).
\end{equation} 

Additionally, we impose constraints on the similarity distribution of the unauthorized domain's text features, ensuring they maintain low entropy through:
\begin{equation}
\label{eq9}
\mathcal{L}_{en} = \mathcal{L}_{entropy}(f_t^u).
\end{equation} 

Finally, our overall loss function can be expressed as:
\begin{equation}
\label{eq10}
\mathcal{L} = \mathcal{L}_{a} - \mathcal{L}_{u} + \mathcal{L}_{ai} - \mathcal{L}_{ui} - \mathcal{L}_{kl} - \lambda_1 \cdot \mathcal{L}_{m} + \lambda_2 \cdot \mathcal{L}_{en}.
\end{equation} 
Where $\lambda_1$ and $\lambda_2$ are weight factors. The overall training strategy is shown in \textit{Supplementary Algorithm 1}.

\textbf{Target-free IP-CLIP.}
In a restricted setting where only authorized domain data is accessible, our IP protection focuses on reducing recognition performance for potential out-of-domain (OOD) data with similar content but different styles. Unlike Wang~\cite{CUTI}'s use of GANs for OOD data synthesis, we intervene on the style factor to achieve this. Our method enhances style~\cite{randaugment} without changing the content (as in \textit{Supplementary Tab.~1}). We treat all style-augmented images as unauthorized and train the model similarly to target-specific IP-CLIP. The full algorithm is detailed in \textit{Supplementary Algorithm 2}.

\textbf{Inference.}
During testing, as shown in~\cref{figure2} (b), the sample is input into visual encoder, and the trained IP Projector generates the corresponding prompt, which is then fed into text encoder. Finally, the cosine similarity between $f_v$ and $f_t$ is computed to produce the prediction $p$:
\begin{equation}
\label{pred}
p = \arg\max_{i} \langle f_t, f_{v,i} \rangle,
\end{equation} 
where $i$ denote the index of class.

%% file: sec/4_exp.tex
\section{Experiment}
\label{sec:exp}

\subsection{Implementation Details}

We evaluated our method on three popular domain adaptation / generalization benchmarks, which feature more categories, larger numbers, and more complex content compared to the existing works~~\cite{NTL,CUTI, CUPI}:
\begin{enumerate}
\item \textbf{Office-31~\cite{office31}} comprises images from three distinct domains—Amazon, Dslr, and Webcam—spanning 31 categories and containing over 4,000 samples.
\item \textbf{Office-Home-65~\cite{home}} consists of over 15,000 images distributed across four domains—Art, Clipart, Product, and Real-World—organized into 65 distinct categories.
\item \textbf{Mini-DomainNet~\cite{mini}} contains over 140,000 images across domains including Clipart, Painting, Real, and Sketch, with 126 categories.
\end{enumerate}
The substantial differences in image style and quality across domains in these datasets make them ideal for evaluating the effectiveness of model IP protection algorithms in cross-domain image recognition tasks.

Our comprehensive experiments are implemented on the PyTorch platform and an NVIDIA GeForce RTX 3090 GPU with 24GB of memory. The Adam optimizer, with an initial learning rate of $e^{-5}$, is employed for model optimization. We utilize the pre-trained CLIP backbone architecture. Consistent with standard evaluation protocols, accuracy (\%) is used as the primary performance metric for each task.

\subsection{Result of Target-Specified IP-CLIP}
In the target-specified scenario, we randomly select two domains from each dataset: one as the authorized domain and the other as the unauthorized domain, thereby forming a IP protection task. We first compute $A_a^{SL}/A_u^{SL}$, the performance of supervised learning CLIP with prompt fine-tuning (SL-CLIP) trained on the authorized domain and tested on the authorized / unauthorized domain, and $A_a^{IP}/A_u^{IP}$, the performance of IP-CLIP on the same domain. This process is denoted as: $A^{SL} \Rightarrow A^{IP}$, with results shown in~\cref{ts_office31_detail}. Given CLIP's strong feature extraction capabilities, it tends to generalize well, resulting in higher $A^{SL}$. However, our goal is to restrict the model to the authorized domain, leading to a lower $A^{IP}$. Additionally, the previous method only assessed the drop rates $D_a=A^{SL}_a-A^{IP}_a$ for the authorized and $D_u=\mu (A^{SL}_u-A^{IP}_u)$ for the unauthorized domains, which is insufficient. An effective IP protection model must balance maintaining high performance in the authorized domain with degrading performance in the unauthorized domain. To address this, we define a new weighted metric, $W_{ua}$, as follows:
\begin{equation}
\label{eq11}
W_{ua} = A^{IP}_a \cdot [D_u - D_a].
\end{equation} 

\cref{ts_office31_compare} present the performance comparison between the proposed IP-CLIP and SOTA methods on the Office-31~\cite{office31}. The results for CUTI~\cite{CUTI} and NTL~\cite{NTL} were obtained by reproducing their original implementations. For a fair comparison, we adapted these methods into CLIP-based versions, referred to as CUTI$^{\dagger}$ and NTL$^{\dagger}$, respectively. The results indicate that the CLIP-based model exhibits stronger protection capabilities compared to the CNN-based model, achieving an average $W_{ua}$ of 74.84\% for IP-CLIP, 72.48\% for CUTI$^{\dagger}$, 54.98\% for NTL$^{\dagger}$, 70.09\% for CUTI, and 62.11\% for NTL. IP-CLIP achieves the highest scores across nearly all metrics. Although CUTI slightly outperforms IP-CLIP in $D_{u}$ in the "webcam" domain, its $D_{a}$ is 2.5\%, significantly above IP-CLIP's 0.0\%. The goal of the IP protection task is to reduce performance in the unauthorized domain while preserving accuracy in the authorized domain. Thus, relying solely on $D_{u}$ or $D_{a}$ is insufficient for comprehensive evaluation, making a combined metric like $W_{ua}$ essential for a balanced assessment.

\begin{table}[!t]
\renewcommand\arraystretch{0.9}
  \centering
  \resizebox{0.48\textwidth}{!}{
  \begin{tabular}{c|ccc|ccc}
    \toprule
    Authorized/Unauthorized & Amazon & Dslr & Webcam & $W_{ua} \uparrow$ & $D_u \uparrow$ & $D_a \downarrow$ \\
    \midrule   
    Amazon & 79.4 $\Rightarrow$  79.4 & 87.5 $\Rightarrow$ \ 7.5 & 88.8 $\Rightarrow$ \ 8.8 & 63.52 & 80.00 & 0.00 \\ 
    Dslr   & 83.8 $\Rightarrow$ \ 3.8 & 95.7 $\Rightarrow$  95.7 & 98.8 $\Rightarrow$ \ 6.3 & 82.54 & 86.25 & 0.00 \\ 
    Webcam & 80.0 $\Rightarrow$ \ 3.8 & 92.5 $\Rightarrow$ \ 2.5 & 94.4 $\Rightarrow$  94.4 & 78.45 & 83.10 & 0.00 \\
    \midrule
    Mean   & \multicolumn{3}{c|}{/} & 74.84 & 83.12 & 0.00 \\
    \bottomrule
  \end{tabular}}
    \caption{The accuracy ($\%$) of target-specified IP-CLIP on the Office-31~\cite{office31}. The vertical/horizontal axis denotes the authorized/unauthorized domain. In each task, the left of '\(\Rightarrow\)' shows the test accuracy of supervised learning CLIP on the unauthorized domain, while the right side presents the accuracy of IP-CLIP. $W_{ua}$ represents the proposed weighted drop, while $D_u$ and $D_a$ denote the drop rates for the unauthorized and authorized domains, respectively.}
  \label{ts_office31_detail}
\end{table}

\begin{table*}[!t]
\renewcommand\arraystretch{1}
\setlength{\tabcolsep}{2pt}
  \centering
  \resizebox{1\textwidth}{!}{
  \begin{tabular}{c|c|ccccc|ccccc|ccccc}
    \toprule
    \multirow{2}{*}{Datasets} & \multirow{2}{*}{\makecell[c]{Authorized \\ Domain}} & \multicolumn{5}{c|}{$W_{ua} \uparrow$} & \multicolumn{5}{c|}{$D_u \uparrow$} & \multicolumn{5}{c}{$D_a \downarrow$} \\
    & & NTL~\cite{NTL} & CUTI~\cite{CUTI} & NTL$^{\dagger}$~\cite{NTL} & CUTI$^{\dagger}$~\cite{CUTI} & IP-CLIP & NTL~\cite{NTL} & CUTI~\cite{CUTI} & NTL$^{\dagger}$~\cite{NTL} & CUTI$^{\dagger}$~\cite{CUTI} & IP-CLIP & NTL~\cite{NTL} & CUTI~\cite{CUTI} & NTL$^{\dagger}$~\cite{NTL} & CUTI$^{\dagger}$~\cite{CUTI} & IP-CLIP \\
    \midrule 
    \multirow{4}{*}{\makecell[c]{Office-31 \\~\cite{office31}}} 
    & Amazon & 41.37 & 60.94 & 56.34 & 62.06 & \bf 63.52 & 55.50 & 74.40 & 75.80 &     79.35 & \bf 80.00 &     3.10 &     0.80 & 1.80 & 0.60 & \bf 0.00 \\
    & Dslr   & 70.94 & 75.33 & 76.09 & 80.13 & \bf 82.54 & 74.20 & 81.90 & 77.35 &     85.05 & \bf 86.25 &     1.55 &     0.80 & 1.30 & 0.70 & \bf 0.00 \\
    & Webcam & 74.02 & 74.02 & 32.50 & 75.24 & \bf 78.45 & 75.80 & 38.70 & 75.80 & \bf 84.38 &     83.10 & \bf 0.00 & \bf 0.00 & 3.10 & 2.50 & \bf 0.00 \\
    & \cellcolor{gray!20} Mean   & \cellcolor{gray!20} 62.11 & \cellcolor{gray!20} 70.09 & \cellcolor{gray!20} 54.98 & \cellcolor{gray!20} 72.48 & \cellcolor{gray!20} \bf 74.84$^{\ast}$ & \cellcolor{gray!20} 68.50 & \cellcolor{gray!20} 76.32 & \cellcolor{gray!20} 65.00 & \cellcolor{gray!20} 82.93 & \cellcolor{gray!20} \bf 83.12 & \cellcolor{gray!20} 1.55 & \cellcolor{gray!20} 0.53 & \cellcolor{gray!20} 2.07 & \cellcolor{gray!20} 1.27 & \cellcolor{gray!20} \bf 0.00$^{\ast}$ \\
    \midrule                  
    \multirow{5}{*}{\makecell[c]{Office- \\ Home-65 \\~\cite{home}}}
    & Art       & 27.53 & 35.62 & 13.44 & 41.58 & \bf 52.00 & 37.27 & 47.16 & 15.83 & 53.40 & \bf 61.33 & 0.80 & 0.30 & \bf 0.10 &     3.00 &     0.30 \\
    & Clipart   & 43.23 & 45.67 & 48.83 & 53.37 & \bf 56.45 & 54.31 & 57.35 & 65.67 & 72.40 & \bf 75.47 & 0.20 & 0.20 &     0.30 &     0.63 & \bf 0.10 \\
    & Product   & 41.31 & 41.78 & 39.90 & 56.82 & \bf 58.71 & 45.01 & 45.82 & 43.00 & 61.83 & \bf 63.77 & 0.30 & 0.50 & \bf 0.00 &     0.37 &     0.30 \\
    & RealWorld & 22.93 & 35.87 & 28.87 & 49.41 & \bf 53.25 & 30.37 & 42.95 & 34.67 & 57.33 & \bf 59.33 & 2.40 & 0.30 &     1.90 &     1.50 & \bf 0.10 \\
    & \cellcolor{gray!20} Mean      & \cellcolor{gray!20} 33.75 & \cellcolor{gray!20} 39.73 & \cellcolor{gray!20} 32.76 & \cellcolor{gray!20} 50.29 & \cellcolor{gray!20} \bf 55.10$^{\ast}$ & \cellcolor{gray!20} 41.74 & \cellcolor{gray!20} 48.32 & \cellcolor{gray!20} 39.79 & \cellcolor{gray!20} 61.24 & \cellcolor{gray!20} \bf 64.98$^{\ast}$ & \cellcolor{gray!20} 0.43 & \cellcolor{gray!20} 0.33 & \cellcolor{gray!20} 0.57 & \cellcolor{gray!20} 1.38 & \cellcolor{gray!20} \bf 0.20 \\
    \midrule                  
    \multirow{5}{*}{\makecell[c]{Mini- \\ DomainNet \\~\cite{mini}}}  
    & Clipart   & 25.63 & 30.29 & 38.62 & 50.26 & \bf 51.47 & 36.60 & 40.87 & 46.30 & 59.40 & \bf 61.00 &     2.10 & 0.80 & 0.60 & \bf 0.20 &     0.30 \\ 
    & Painting  & 19.53 & 19.88 & 41.66 & 46.88 & \bf 53.85 & 32.37 & 33.23 & 53.80 & 66.90 & \bf 67.07 & \bf 0.50 & 0.70 & 1.60 &     5.30 & \bf 0.50 \\ 
    & Real      & 29.26 & 31.52 & 52.29 & 54.77 & \bf 58.82 & 35.87 & 38.40 & 59.03 & 62.30 & \bf 65.27 &     1.20 & 1.10 & 0.80 &     1.10 & \bf 0.20 \\
    & Sketch    & 29.37 & 30.18 & 33.78 & 51.09 & \bf 54.59 & 45.77 & 46.90 & 42.77 & 64.57 & \bf 68.57 &     1.00 & 0.96 & 0.60 &     0.70 & \bf 0.50 \\
    & \cellcolor{gray!20} Mean      & \cellcolor{gray!20} 25.95 & \cellcolor{gray!20} 27.97 & \cellcolor{gray!20} 41.59 & \cellcolor{gray!20} 50.75 & \cellcolor{gray!20} \bf 54.68$^{\ast}$ & \cellcolor{gray!20} 37.65 & \cellcolor{gray!20} 39.85 & \cellcolor{gray!20} 50.48 & \cellcolor{gray!20} 63.29 & \cellcolor{gray!20} \bf 65.48$^{\ast}$ & \cellcolor{gray!20} 1.27 & \cellcolor{gray!20} 0.87 & \cellcolor{gray!20} 1.00 & \cellcolor{gray!20} 2.20 & \cellcolor{gray!20} \bf 0.33$^{\ast}$ \\
    \bottomrule
  \end{tabular}}
    \caption{$W_{ua}$, $D_u$, and $D_a$ of target-specified IP-CLIP, CUTI$^{\dagger}$, NTL$^{\dagger}$, CUTI and NTL. $W_{ua}$ represents the proposed weighted drop, while $D_u$ and $D_a$ denote the drop rates for the unauthorized and authorized domains, respectively. The best performance is indicated by the numbers in bold. Statistical significance (p-value $<$ 0.05~\cite{p1,p2}) is denoted with: $^{\ast}$(IP-CLIP vs. others).}
  \label{ts_office31_compare}
\end{table*}

\begin{figure}[t]
  \centering
   \includegraphics[width=0.7\linewidth,trim=100 205 460 205,clip]{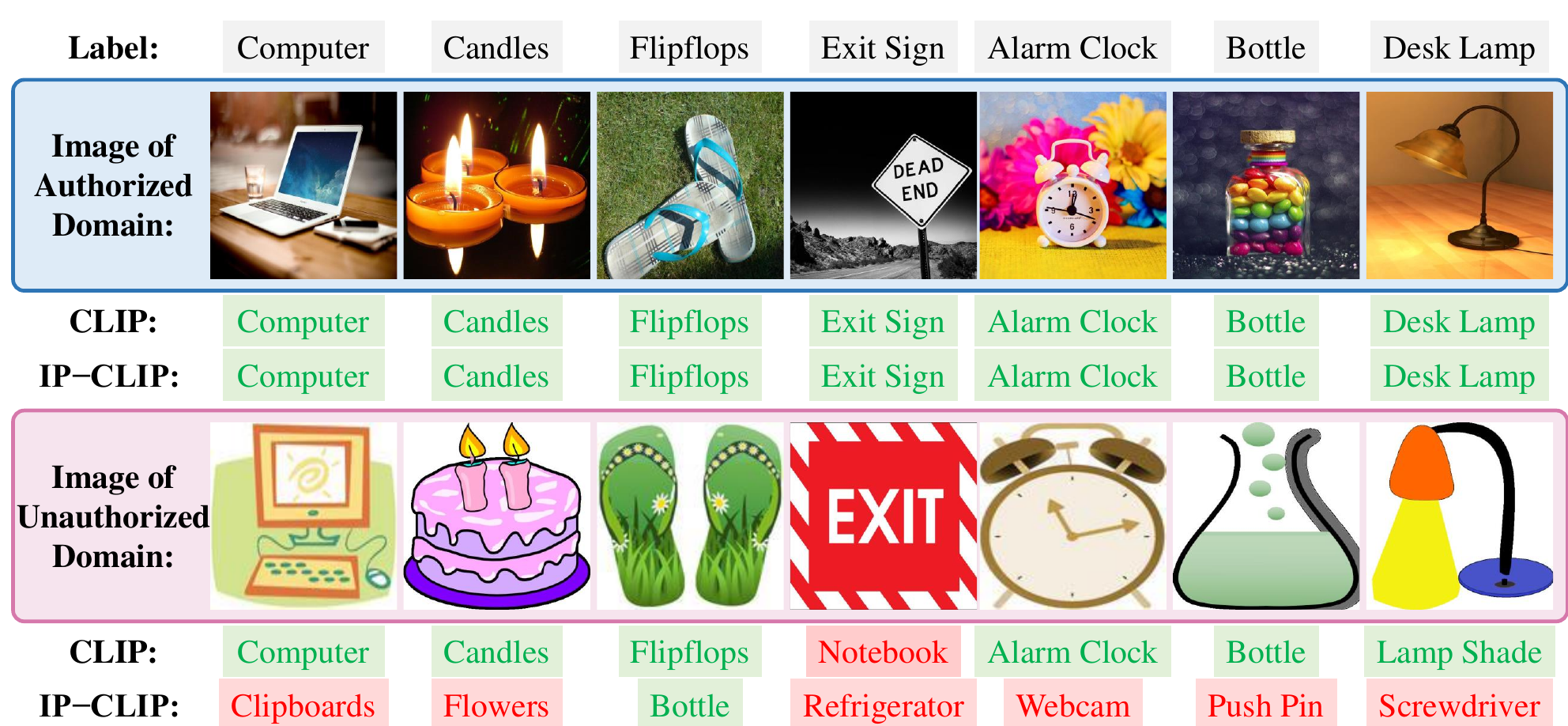}
   \caption{Several visualization examples of CLIP and IP-CLIP prediction results. Correct predictions are highlighted in green, while incorrect predictions are shown in red.} 
   \label{figure3}
\end{figure}

Additionally, we evaluated the proposed IP-CLIP on Office-Home-65~\cite{home} and Mini-DomainNet~\cite{mini} to further verify its effectiveness and versatility. The experimental results are summarized in~\cref{ts_office31_compare}, with further details available in \textit{Supplementary Tab.~2-16}. Across these datasets, the CLIP-based IP protection scheme consistently outperforms its CNN counterpart, with IP-CLIP demonstrating the strongest protection capabilities.~\cref{figure3} presents several visualization examples.

\subsection{Result of Ownership Verification}\label{ov}

\begin{table*}[!t]
\renewcommand\arraystretch{1}
\setlength{\tabcolsep}{2pt}
  \centering
  \resizebox{1\textwidth}{!}{
  \begin{tabular}{c|c|c|cc|cc|c|cc|cc|cc}
    \toprule

    \multirow{3}{*}{Datasets} & \multirow{3}{*}{\makecell[c]{Authorized \\ with / without \\ Patch}} & \multicolumn{5}{c|}{CNN-Based Models} & \multicolumn{7}{c}{CLIP-Based Modesl} \\
    \cline{3-14}
    & & SL-CNN & \multicolumn{2}{c|}{NTL~\cite{NTL}} & \multicolumn{2}{c|}{CUTI~\cite{CUTI}} & SL-CLIP~\cite{clip} & \multicolumn{2}{c|}{NTL$^{\dagger}$~\cite{NTL}} & \multicolumn{2}{c|}{CUTI$^{\dagger}$~\cite{CUTI}} & \multicolumn{2}{c}{IP-CLIP} \\
    & & $A_u/A_a$ & $A_u/A_a$ & \cellcolor{gray!20} $O_{ua} \uparrow$ & $A_u/A_a$ & \cellcolor{gray!20} $O_{ua} \uparrow$ & $A_u/A_a$ & $A_u/A_a$ & \cellcolor{gray!20} $O_{ua} \uparrow$ & $A_u/A_a$ & \cellcolor{gray!20} $O_{ua} \uparrow$ & $A_u/A_a$ & \cellcolor{gray!20} $O_{ua} \uparrow$ \\
    \hline                            
    \multirow{3}{*}{\makecell[c]{Office-31 \\~\cite{office31}}}     
    & Amazon    & 59.4 $/$ 78.1 & 3.1 $/$ 67.2 & \cellcolor{gray!20} 38.1 & 1.6 $/$ 78.1 & \cellcolor{gray!20} 45.4 & 80.0 $/$ 81.3 & 15.0 $/$ 77.5 & \cellcolor{gray!20} 50.0 &  3.8 $/$ 80.0 & \cellcolor{gray!20} 61.0 & 3.8 $/$ 81.3 & \cellcolor{gray!20} \bf 62.0 \\
    & Dslr      & 50.0 $/$ 98.4 & 0.0 $/$ 92.2 & \cellcolor{gray!20} 46.1 & 4.7 $/$ 93.8 & \cellcolor{gray!20} 44.6 & 97.5 $/$ 98.8 &  5.0 $/$ 95.0 & \cellcolor{gray!20} 87.8 &  2.5 $/$ 95.0 & \cellcolor{gray!20} 90.2 & 3.8 $/$ 97.5 & \cellcolor{gray!20} \bf 91.4 \\
    & Webcam    & 62.5 $/$ 95.3 & 1.6 $/$ 93.8 & \cellcolor{gray!20} 57.6 & 4.7 $/$ 92.2 & \cellcolor{gray!20} 54.7 & 95.0 $/$ 97.5 &  2.5 $/$ 93.8 & \cellcolor{gray!20} 86.7 &  7.5 $/$ 95.0 & \cellcolor{gray!20} 83.1 & 1.3 $/$ 96.3 & \cellcolor{gray!20} \bf 90.3 \\
    \hline    
    \multirow{4}{*}{\makecell[c]{Office- \\ Home-65 \\~\cite{home}}}             
    & Art       & 54.7 $/$ 76.8 & 1.6 $/$ 45.6 & \cellcolor{gray!20} 24.1 & 1.6 $/$ 76.0 & \cellcolor{gray!20} 40.7 & 83.5 $/$ 85.5 & 16.5 $/$ 87.3 & \cellcolor{gray!20} 59.1 &  6.0 $/$ 87.0 & \cellcolor{gray!20} 67.6 & 5.0 $/$ 87.5 & \cellcolor{gray!20} \bf 68.9 \\
    & Clipart   & 70.8 $/$ 78.1 & 1.6 $/$ 54.9 & \cellcolor{gray!20} 37.7 & 3.1 $/$ 69.0 & \cellcolor{gray!20} 46.7 & 73.8 $/$ 74.3 &  5.5 $/$ 73.5 & \cellcolor{gray!20} 50.2 & 17.0 $/$ 73.3 & \cellcolor{gray!20} 41.5 & 5.5 $/$ 73.5 & \cellcolor{gray!20} \bf 50.2 \\
    & Product   & 65.9 $/$ 92.2 & 2.3 $/$ 69.8 & \cellcolor{gray!20} 44.5 & 2.6 $/$ 91.1 & \cellcolor{gray!20} 58.3 & 90.5 $/$ 94.0 & 60.5 $/$ 92.5 & \cellcolor{gray!20} 29.0 & 31.0 $/$ 93.0 & \cellcolor{gray!20} 56.1 & 2.0 $/$ 92.8 & \cellcolor{gray!20} \bf 82.2 \\
    & RealWorld & 61.2 $/$ 82.6 & 1.8 $/$ 77.3 & \cellcolor{gray!20} 46.2 & 0.3 $/$ 83.6 & \cellcolor{gray!20} 51.0 & 87.5 $/$ 88.5 & 17.5 $/$ 87.8 & \cellcolor{gray!20} 61.5 &  5.0 $/$ 86.3 & \cellcolor{gray!20} 71.1 & 6.5 $/$ 92.0 & \cellcolor{gray!20} \bf 74.8 \\
    \hline                            
    \multirow{4}{*}{\makecell[c]{Mini- \\ DomainNet \\~\cite{mini}}}     
    & Clipart   & 50.3 $/$ 65.5 & 0.8 $/$ 37.8 & \cellcolor{gray!20} 18.6 & 1.6 $/$ 67.8 & \cellcolor{gray!20} 33.3 & 84.0 $/$ 85.1 & 57.1 $/$ 86.4 & \cellcolor{gray!20} 24.6 & 13.7 $/$ 85.2 & \cellcolor{gray!20} 60.1 & 5.6 $/$ 85.4 & \cellcolor{gray!20} \bf 67.0 \\
    & Painting  & 39.6 $/$ 57.6 & 0.8 $/$ 46.1 & \cellcolor{gray!20} 17.9 & 1.0 $/$ 56.9 & \cellcolor{gray!20} 22.1 & 79.5 $/$ 81.9 & 31.1 $/$ 80.0 & \cellcolor{gray!20} 38.9 &  4.1 $/$ 78.8 & \cellcolor{gray!20} 59.4 & 4.1 $/$ 81.1 & \cellcolor{gray!20} \bf 61.2 \\
    & Real      & 50.2 $/$ 82.6 & 0.0 $/$ 40.3 & \cellcolor{gray!20} 20.2 & 0.5 $/$ 83.2 & \cellcolor{gray!20} 41.5 & 88.9 $/$ 89.4 & 26.2 $/$ 91.9 & \cellcolor{gray!20} 58.4 & 11.4 $/$ 92.1 & \cellcolor{gray!20} 71.7 & 5.9 $/$ 89.7 & \cellcolor{gray!20} \bf 74.5 \\
    & Sketch    & 57.6 $/$ 63.5 & 0.3 $/$ 57.4 & \cellcolor{gray!20} 32.9 & 0.7 $/$ 61.3 & \cellcolor{gray!20} 34.9 & 81.0 $/$ 81.0 & 39.7 $/$ 79.7 & \cellcolor{gray!20} 32.4 &  4.8 $/$ 79.7 & \cellcolor{gray!20} 60.7 & 2.5 $/$ 79.1 & \cellcolor{gray!20} \bf 62.0 \\
    \hline                            
    \multicolumn{2}{c|}{Mean} & / & / & \cellcolor{gray!20} 34.9 & / & \cellcolor{gray!20} 43.0 & / & / & \cellcolor{gray!20} 52.6 & / & \cellcolor{gray!20} 65.7 & / & \cellcolor{gray!20} \bf 71.3$^{\ast}$ \\
    \bottomrule
  \end{tabular}}
    \caption{The results of ownership verification by SL-CNN~\cite{CUTI}, NTL~\cite{NTL}, CUTI~\cite{CUTI}, NTL$^{\dagger}$, CUTI$^{\dagger}$, and IP-CLIP. $O_{ua}$ represents the proposed weighted drop, while $A_u$ and $A_a$ denote the accuarcy for the domain with and without patch, respectively. The best performance is indicated by the numbers in bold. Statistical significance (p-value $<$ 0.05~\cite{p1,p2}) is denoted with: $^{\ast}$(IP-CLIP vs. others).}
  \label{ov_detail}
\end{table*}

To further verify model ownership, erroneous results are deliberately triggered. Specifically, a conventional backdoor watermark is applied to each authorized domain~\cite{CUTI}, with the processed data used as the corresponding unauthorized domain. For ease of observation and analysis, we computed the accuracy of the supervised convolutional neural network (SL-CNN) related to CNN-based NTL/CUTI, as well as the supervised CLIP (SL-CLIP) according to CLIP-based NTL$^{\dagger}$/CUTI$^{\dagger}$/IP-CLIP. After computing $A_a$ and $A_u$, a new weighted metric is introduced based on these values: 
\begin{equation}
\label{eq12}
O_{ua} = A_u^{SL} \cdot [A_a^{Method} - A_u^{Method}].
\end{equation} 

As presented in~\cref{ov_detail}, the difference in accuracy between SL-CNN/SL-CLIP with a watermark ($A_a^{SL}$) and without a watermark ($A_u^{SL}$) is minimal, indicating low sensitivity to the watermark. In contrast, IP-CLIP shows a significant reduction in accuracy on unauthorized domains with embedded watermarks ($A_u^{IP}$). This disparity in performance serves as an effective measure for verifying model ownership. Furthermore, the performance comparison between IP-CLIP and other state-of-the-art methods reveals that, compared to CNN-based models, CLIP-based models show stronger model protection capabilities. Notably, $O_{ua}$ of IP-CLIP is 71.3\%, outperforming CUTI$^{\dagger}$ and NTL$^{\dagger}$ by approximately 5.6\% and 18.7\%, respectively, with statistically significant differences (p $<$ 0.05~\cite{p1,p2}).

\subsection{Result of Target-Free IP-CLIP}

\begin{table}[!t]
\renewcommand\arraystretch{0.9}
  \centering
  \resizebox{0.48\textwidth}{!}{
  \begin{tabular}{c|ccc|ccc}
    \toprule
    Authorized/Test & Amazon & Dslr & Webcam & $W_{ua} \uparrow$ & $D_u \uparrow$ & $D_a \downarrow$ \\
    \midrule   
    Amazon & 79.4 $\Rightarrow$ 79.0 & 87.5 $\Rightarrow$ \ 9.8 & 88.8 $\Rightarrow$ 38.3 & 50.32 & 64.10 & 0.40 \\ 
    Dslr   & 83.8 $\Rightarrow$ 23.3 & 95.7 $\Rightarrow$  95.3 & 98.8 $\Rightarrow$ 64.3 & 44.89 & 47.50 & 0.40 \\ 
    Webcam & 80.0 $\Rightarrow$ 17.8 & 92.5 $\Rightarrow$  10.0 & 94.4 $\Rightarrow$ 92.5 & 65.17 & 72.35 & 1.90 \\
    \midrule
    Mean   & \multicolumn{3}{c|}{/} & 53.46 & 61.32 & 0.90 \\
    \bottomrule
  \end{tabular}}
    \caption{The accuracy ($\%$) of target-free IP-CLIP on the Office-31~\cite{office31}. The vertical/horizontal axis denotes the authorized/test domain.}
  \label{tf_office31_detail}
\end{table}

\begin{table*}[!t]
\renewcommand\arraystretch{1}
\setlength{\tabcolsep}{2pt}
  \centering
  \resizebox{1\textwidth}{!}{
  \begin{tabular}{c|c|ccccc|ccccc|ccccc}
    \toprule
    \multirow{2}{*}{Datasets} & \multirow{2}{*}{\makecell[c]{Authorized \\ Domain}} & \multicolumn{5}{c|}{$W_{ua} \uparrow$} & \multicolumn{5}{c|}{$D_u \uparrow$} & \multicolumn{5}{c}{$D_a \downarrow$} \\
    & & NTL~\cite{NTL} & CUTI~\cite{CUTI} & NTL$^{\dagger}$~\cite{NTL} & CUTI$^{\dagger}$~\cite{CUTI} & IP-CLIP & NTL~\cite{NTL} & CUTI~\cite{CUTI} & NTL$^{\dagger}$~\cite{NTL} & CUTI$^{\dagger}$~\cite{CUTI} & IP-CLIP & NTL~\cite{NTL} & CUTI~\cite{CUTI} & NTL$^{\dagger}$~\cite{NTL} & CUTI$^{\dagger}$~\cite{CUTI} & IP-CLIP \\
    \midrule 
    \multirow{4}{*}{\makecell[c]{Office-31 \\~\cite{office31}}} 
    & Amazon & 0.56 & 4.69 & 11.90 & 25.60 & \bf 50.32 & 7.80 &  13.30 & 17.25 & 36.65 & \bf 64.10 & 7.05 & 7.05 & 1.90 &     3.10 & \bf 0.40 \\ 
    & Dslr   & 6.88 & 6.83 & 36.72 & 38.83 & \bf 44.89 & 9.40 & \ 9.35 & 43.90 & 43.30 & \bf 47.50 & 2.30 & 2.30 & 3.90 &     1.90 & \bf 0.40 \\
    & Webcam & 2.90 & 2.95 & 45.80 & 30.95 & \bf 65.17 & 8.60 & \ 5.45 & 50.95 & 33.60 & \bf 72.35 & 5.45 & 2.35 & 1.60 & \bf 0.60 &     1.90 \\
    & \cellcolor{gray!20} Mean & \cellcolor{gray!20} 3.45 & \cellcolor{gray!20} 4.82 & \cellcolor{gray!20} 31.47 & \cellcolor{gray!20} 31.80 & \cellcolor{gray!20} \bf 53.46$^{\ast}$ & \cellcolor{gray!20} 8.60 & \cellcolor{gray!20} 9.37 & \cellcolor{gray!20} 37.37 & \cellcolor{gray!20} 37.85 & \cellcolor{gray!20} \bf 61.32$^{\ast}$ & \cellcolor{gray!20} 4.93 & \cellcolor{gray!20} 3.90 & \cellcolor{gray!20} 2.47 & \cellcolor{gray!20} 1.87 & \cellcolor{gray!20} \bf 0.90 \\
    \midrule                  
    \multirow{5}{*}{\makecell[c]{Office- \\ Home-65 \\~\cite{home}}}
    & Art       & 0.10 & -0.19 & -0.71 &  -0.65 & \bf \ 4.82 & 1.93 & \ 6.53 & \ 2.83 & \ 3.40 & \bf 12.07 & \bf 1.80 & 6.80 & 3.70 & 4.20 &     6.00 \\ 
    & Clipart   & 0.75 &  1.36 &  0.30 &   5.19 & \bf  14.88 & 1.34 & \ 8.24 & \ 0.90 & \ 8.23 & \bf 19.83 &     0.40 & 6.40 & 0.50 & 1.20 & \bf 0.00 \\ 
    & Product   & 3.13 &  4.21 & 14.08 &  12.57 & \bf  23.67 & 6.08 &  13.08 &  19.03 &  18.50 & \bf 30.40 & \bf 2.60 & 8.10 & 3.30 & 4.30 &     3.80 \\
    & RealWorld & 2.39 &  3.72 & 13.07 &   3.82 & \bf  20.41 & 2.83 & \ 8.83 &  17.67 & \ 5.50 & \bf 22.93 & \bf 0.00 & 4.20 & 2.70 & 1.20 &     0.20 \\
    & \cellcolor{gray!20} Mean & \cellcolor{gray!20} 1.59 & \cellcolor{gray!20} 2.28 & \cellcolor{gray!20} 6.68 & \cellcolor{gray!20} 5.23 & \cellcolor{gray!20} \bf  15.95$^{\ast}$ & \cellcolor{gray!20} 3.05 & \cellcolor{gray!20} \ 9.17 & \cellcolor{gray!20} 10.11 & \cellcolor{gray!20} \ 8.91 & \cellcolor{gray!20} \bf 21.31$^{\ast}$ & \cellcolor{gray!20} \bf 1.20 & \cellcolor{gray!20} 6.38 & \cellcolor{gray!20} 2.55 & \cellcolor{gray!20} 2.73 & \cellcolor{gray!20} 2.50 \\
    \midrule                  
    \multirow{5}{*}{\makecell[c]{Mini- \\ DomainNet \\~\cite{mini}}}  
    & Clipart   & -3.25 & -1.85 & -0.89 & 2.24 & \bf \ 2.95 &      11.80 & 5.30 &     3.50 & 7.07 & \bf  7.63 &  17.30 & 8.00 &     4.60 &     4.30 & \bf 4.00 \\ 
    & Painting  & -0.52 &  0.27 &  0.39 & 0.21 & \bf \ 0.97 &     \ 7.53 & 3.87 & \bf 4.40 & 3.57 &      3.93 & \ 8.50 & 3.40 &     3.90 &     3.30 & \bf 2.70 \\ 
    & Real      &  2.60 &  2.05 &  4.46 & 5.86 & \bf  13.77 &     \ 5.73 & 6.00 &     9.37 & 8.93 & \bf 18.13 & \ 2.60 & 3.50 &     4.20 & \bf 2.30 &     2.50 \\
    & Sketch    &  2.44 & -1.63 &  3.07 & 1.29 & \bf \ 3.74 & \bf  14.53 & 6.70 &     7.37 & 5.17 &      8.23 &  10.20 & 9.56 & \bf 3.40 &     3.50 & \bf 3.40 \\
    & \cellcolor{gray!20} Mean & \cellcolor{gray!20} 0.32 & \cellcolor{gray!20} -0.29 & \cellcolor{gray!20} 1.76 & \cellcolor{gray!20} 2.40 & \cellcolor{gray!20} \bf \ 5.36$^{\ast}$ & \cellcolor{gray!20} \bf \ 9.90 & \cellcolor{gray!20} 5.47 & \cellcolor{gray!20} 6.16 & \cellcolor{gray!20} 6.18 & \cellcolor{gray!20} 9.48 & \cellcolor{gray!20} \ 9.47 & \cellcolor{gray!20} 4.97 & \cellcolor{gray!20} 4.23 &  \cellcolor{gray!20}3.30 & \cellcolor{gray!20} \bf 3.07$^{\ast}$ \\ 
    \bottomrule
  \end{tabular}}
    \caption{$W_{ua}$, $D_u$, and $D_a$ of target-free IP-CLIP, CUTI$^{\dagger}$, NTL$^{\dagger}$, CUTI and NTL. $W_{ua}$ represents the proposed weighted drop, while $D_u$ and $D_a$ denote the drop rates for the unauthorized and authorized domains, respectively. The best performance is indicated by the numbers in bold. Statistical significance (p-value $<$ 0.05~\cite{p1,p2}) is denoted with: $^{\ast}$(IP-CLIP vs. others).}
  \label{tf_office31_compare}
\end{table*}

In a more rigorous setting, i.e., the target-free scenario, we generate unauthorized domains for each authorized domain, as described in~\cref{training}. Specifically, to assess the performance of target-free IP-CLIP on the Office-31~\cite{office31} dataset, we conduct three transfer tasks. For each task, one domain is selected as the authorized domain, with unauthorized domains generated accordingly, while the remaining unknown domains are used for testing. The experimental results are presented in~\cref{tf_office31_detail} and~\cref{tf_office31_compare}.

Similarly, we constructed tasks using more  datasets and compared the results with the SOTA method, as shown in~\cref{tf_office31_compare} (with additional details provided in \textit{Supplementary Tab.~17-31}). After analyzing the results, we found that IP-CLIP consistently achieved the highest $W_{ua}$ across all three datasets. This demonstrates its ability to effectively reduce recognition accuracy for unauthorized domains while maintaining strong recognition performance for authorized domains, even in tasks of varying complexity, thus proving its effectiveness in the restricted model IP protection task.

\subsection{Result of Applicability Authorization}

\begin{table*}[!t]
\renewcommand\arraystretch{1}
\setlength{\tabcolsep}{2pt}
  \centering
  \resizebox{1\textwidth}{!}{
  \begin{tabular}{c|c|ccccc|ccccc|ccccc}
    \toprule
    \multirow{2}{*}{Dataset} & \multirow{2}{*}{\makecell[c]{Authorized \\ Domain}} & \multicolumn{5}{c|}{$D_{ua} \uparrow$} & \multicolumn{5}{c|}{$A_u \downarrow$} & \multicolumn{5}{c}{$A_a \uparrow$} \\
    & & NTL~\cite{NTL} & CUTI~\cite{CUTI} & NTL$^{\dagger}$~\cite{NTL} & CUTI$^{\dagger}$~\cite{CUTI} & IP-CLIP & NTL~\cite{NTL} & CUTI~\cite{CUTI} & NTL$^{\dagger}$~\cite{NTL} & CUTI$^{\dagger}$~\cite{CUTI} & IP-CLIP & NTL~\cite{NTL} & CUTI~\cite{CUTI} & NTL$^{\dagger}$~\cite{NTL} & CUTI$^{\dagger}$~\cite{CUTI} & IP-CLIP \\
    \midrule 
    \multirow{4}{*}{\makecell[c]{Office-31 \\~\cite{office31}}} 
    & Amazon & \ 1.63 & 27.95 & 15.67 & 29.26 & \bf 37.46 &     5.21 & \bf \ 0.52 & 37.43 & 20.83 & \ 3.53 & 15.63 & 53.13 &     62.50 & \bf 65.50 &     63.00 \\
    & Dslr   & \ 9.23 & 72.92 & 39.25 & 54.47 & \bf 82.42 &     4.69 & \bf \ 4.17 & 50.50 & 36.53 & \ 9.77 & 32.81 & 87.50 &     92.80 &     94.30 & \bf 95.80 \\
    & Webcam &  11.82 & 40.01 & 54.59 & 40.56 & \bf 56.45 & \bf 0.00 &      37.00 & 21.30 & 30.60 &  15.53 & 34.38 & 84.40 & \bf 85.30 &     80.80 &     83.30 \\
    & \cellcolor{gray!20} Mean & \cellcolor{gray!20} 7.56 & \cellcolor{gray!20} 46.96 & \cellcolor{gray!20} 36.50 & \cellcolor{gray!20} 41.43 & \cellcolor{gray!20} \bf 58.78$^{\ast}$ & \cellcolor{gray!20} \bf 3.30 & \cellcolor{gray!20} 13.90 & \cellcolor{gray!20} 36.41 & \cellcolor{gray!20} 29.32 & \cellcolor{gray!20} 9.61$^{\ast}$ & \cellcolor{gray!20} 27.60 & \cellcolor{gray!20} 75.01 & \cellcolor{gray!20} 80.20 & \cellcolor{gray!20} 80.20 & \cellcolor{gray!20} \bf 80.70 \\
    \midrule 
    \multirow{5}{*}{\makecell[c]{Office- \\ Home-65 \\~\cite{home}}}
    & Art       & \ 8.75 &  35.25 &  49.47 & 54.95 & \bf 60.12 & 63.93 & \bf \ 1.04 & 20.45 &  10.38 &     \ 3.88 &     75.52 & 59.90 & \bf 81.30 &     79.50 & \bf 79.50 \\
    & Clipart   & \ 4.98 &  14.78 & \ 9.74 & 16.86 & \bf 26.52 & 50.39 & \bf \ 0.72 & 27.70 &  20.88 &      10.48 & \bf 58.85 & 38.80 &     48.00 &     52.80 &     57.00 \\ 
    & Product   &  17.49 &  33.27 &  44.44 & 39.53 & \bf 57.74 & 58.40 & \bf \ 0.78 & 27.48 &  35.38 &     \ 8.40 &     80.21 & 58.07 &     81.80 & \bf 83.00 &     80.30 \\
    & RealWorld &  15.83 & \ 3.15 &  51.50 & 62.87 & \bf 71.17 & 64.97 &      31.32 & 19.20 & \ 7.83 & \bf \ 5.20 &     83.85 & 39.32 &     82.00 &     83.30 & \bf 87.00 \\
    & \cellcolor{gray!20} Mean & \cellcolor{gray!20} 11.76 & \cellcolor{gray!20} 21.61 & \cellcolor{gray!20} 38.79 & \cellcolor{gray!20} 43.55 & \cellcolor{gray!20} \bf 53.89$^{\ast}$ & \cellcolor{gray!20} 59.42 & \cellcolor{gray!20} \ 8.46 & \cellcolor{gray!20} 23.71 & \cellcolor{gray!20} 18.61 & \cellcolor{gray!20} \bf \ 6.99$^{\ast}$ & \cellcolor{gray!20} 74.61 & \cellcolor{gray!20} 49.02 & \cellcolor{gray!20} 73.28 & \cellcolor{gray!20} 74.65 & \cellcolor{gray!20} \bf 75.95$^{\ast}$ \\
    \midrule                  
    \multirow{5}{*}{\makecell[c]{Mini- \\ DomainNet \\~\cite{mini}}}  
    & Clipart   &  11.96 &  13.75 & 38.45 & 22.77 & \bf 50.88 & 58.06 & 60.53 & 17.54 &  44.08 & \bf  7.35 & 74.18 & \bf 78.13 &     71.40 &     74.60 & 75.10 \\
    & Painting  & \ 7.47 & \ 6.47 & 32.78 & 24.48 & \bf 40.33 & 58.26 & 45.15 & 24.18 &  34.73 & \bf 10.93 & 69.08 &     56.58 & \bf 70.60 &     69.80 & 69.20 \\
    & Real      &  21.08 &  22.62 & 35.66 & 33.56 & \bf 54.06 & 57.03 & 58.43 & 37.90 &  34.83 & \bf 18.40 & 82.57 & \bf 85.03 &     81.60 &     77.90 & 83.30 \\
    & Sketch    & \ 7.72 & \ 7.00 & 38.66 & 48.18 & \bf 48.27 & 58.47 & 57.24 & 16.55 & \ 9.45 & \bf  8.20 & 69.57 &     67.60 &     71.00 & \bf 74.30 & 73.70 \\
    & \cellcolor{gray!20} Mean & \cellcolor{gray!20} 12.06 & \cellcolor{gray!20} 12.46 & \cellcolor{gray!20} 36.39 & \cellcolor{gray!20} 32.25 & \cellcolor{gray!20} \bf 48.39 & \cellcolor{gray!20} 57.96 & \cellcolor{gray!20} 55.34 & \cellcolor{gray!20} 24.04 & \cellcolor{gray!20} 30.77 & \cellcolor{gray!20} \bf 11.22 & \cellcolor{gray!20} 73.85 & \cellcolor{gray!20} 71.83 & \cellcolor{gray!20} 73.65 & \cellcolor{gray!20} 74.15 & \cellcolor{gray!20} \bf 75.33 \\
    \bottomrule
  \end{tabular}}
    \caption{$D_{ua}$, $A_u$, and $A_a$ of authorization application IP-CLIP, CUTI$^{\dagger}$, NTL$^{\dagger}$, CUTI and NTL on the Office-31~\cite{office31}.   
    $D_{ua}$ represents the proposed weighted drop, while $A_u$ and $A_a$ denote the accuracy for the unauthorized and authorized domains, respectively. The best performance is indicated by the numbers in bold. Statistical significance (p-value $<$ 0.05~\cite{p1,p2}) is denoted with: $^{\ast}$(IP-CLIP vs. others).}
  \label{aa_office31_compare}
\end{table*}

\begin{table}[!t]
\renewcommand\arraystretch{0.9}
  \centering
  \resizebox{0.48\textwidth}{!}{
  \begin{tabular}{c|ccc|ccc}
    \toprule
    Authorized/Test & Amazon & Dslr & Webcam & $D_{ua} \uparrow$ & $A_u \downarrow$ & $A_a \uparrow$ \\
    \midrule 
    Amazon & \ 4.5 & 3.3 & \ 2.8 & 37.46 & \ 3.53 & 63.00 \\
    Dslr   &  27.3 & 1.5 & \ 0.5 & 82.42 & \ 9.77 & 95.80 \\
    Webcam &  31.0 & 4.3 &  11.3 & 56.45 &  15.53 & 83.30 \\
    \midrule 
    Mean   & \multicolumn{3}{c|}{/} & 58.78 & \ 9.61 & 80.70 \\
    \bottomrule
  \end{tabular}}
    \caption{$D_{ua}$, $A_u$, and $A_a$ of authorization application IP-CLIP on the Office-31~\cite{office31}. The vertical/horizontal axis denotes the authorized/test domain. $D_{ua}$ represents the proposed weighted drop, while $A_u$ and $A_u$ denote the accuarcy of the unauthorized and test domains, respectively.}
  \label{aa_office31_detail}
\end{table}

In the applicability authorization scenario, we assess the model's effectiveness by limiting its generalization ability to the authorized domain. Specifically, following the approach outlined in~\cref{ov}, we designate one domain as the original domain, to which we apply a specific watermark, resulting in the processed data being classified as the authorized domain. The unauthorized domain set is then formed by mixing the original domain, the domain generated from the original domain, and the generated domain with the watermark. During testing, the original domain and other unknown domains are used as the test set.

\cref{aa_office31_compare} and~\cref{aa_office31_detail} present the experimental results of IP-CLIP and SOTA methods on the Office-31~\cite{office31}, while results from additional datasets are shown in~\cref{aa_office31_compare} (see \textit{Supplementary Tab.~32-46} for further details). An interesting pattern emerges from the \cref{aa_office31_detail}: in some domains, the $A_u$ of NTL and CUTI outperform that of IP-CLIP, while their $A_a$ is lower than that of IP-CLIP, and even in extreme cases is only one-third; Conversely, in certain cases, the $A_a$ performance of NTL, CUTI, and IP-CLIP is comparable, but their $A_u$ performance is worse. This demonstrates that relying on a single indicator (i.e., $A_u$ and $A_a$) to assess IP protection is inadequate, highlighting the need for a comprehensive weighted metric $D_{ua} = A_a \cdot [A_a - A_u]$. As expected, IP-CLIP consistently achieves the highest $D_{ua}$ across various domains, confirming that its generalization is effectively constrained to the authorized domain.

%% file: sec/5_conclusion.tex
\section{Conclusion}
Protecting the intellectual property (IP) of visual language models (VLMs) like CLIP is a significant challenge in artificial intelligence. To address this, we propose IP-CLIP, a lightweight, prompt-based strategy that extracts image style and content for domain verification while preventing unauthorized feature transfers. Extensive experiments on cross-domain datasets demonstrate the effectiveness of our lightweight and easy-to-deploy IP-CLIP. Though designed for classification tasks, IP-CLIP can be extended to applications such as detection and image description. Future work will focus on enhancing generalization and adapting IP protection strategies to diverse model architectures. We believe our work will advance research in model IP protection and underscore its practical importance.

%% file: sec/X_suppl.tex
\clearpage
\setcounter{page}{1}

\setcounter{table}{0}

\begin{figure*}[!t]
  \centering
   \includegraphics[width=1\linewidth,trim=70 680 70 70,clip]{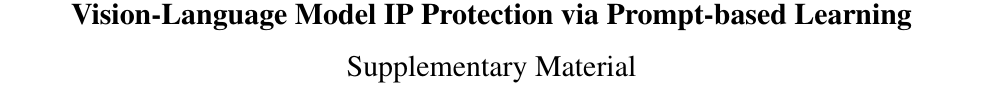}
   \label{r_figure2} 
\end{figure*}

\begin{table*}[!t]
\renewcommand\arraystretch{1.6}
  \centering
  \resizebox{0.6\textwidth}{!}{
  \begin{tabular}{cc}
    \toprule
    Augmentation & Augmentation\\

    \midrule 
    AutoContrast   & Applies automatic contrast adjustment to an image. \\
    Brightness     & Adjusts the brightness of an image. \\
    Color          & Adjusts the color saturation of an image. \\
    Contrast       & Adjusts the contrast of an image. \\
    Equalize       & Equalizes the histogram of an image. \\
    Identity       & Returns the image without any changes. \\
    Posterize      & Reduces the color depth of an image. \\
    Rotate         & Rotates an image by a random degree. \\
    Sharpness      & Adjusts the sharpness of an image. \\
    ShearX         & Shears an image along the X-axis. \\
    ShearY         & Shears an image along the Y-axis. \\
    Solarize       & Inverts all pixel values above a threshold \\
    TranslateX     & Translates an image horizontally. \\
    TranslateY     & Translates an image vertically \\
    \bottomrule
  \end{tabular}}
      \caption{A detailed description of the augmentation method applied in the target-free scenario.}
  \label{r_aug}
\end{table*}

\begin{algorithm*}[!t]
\caption{Target-Specified IP-CLIP.}\label{alg:alg1}
\begin{algorithmic}[1]
\REQUIRE{The authorized domain $D_a$, unauthorized domain $D_u$, visual encoder of CLIP with $L$ layers, text encoder of CLIP, the parameters $\theta$ of IP Projector $P$ and $\phi$ of STAM.}
\STATE $ \text{Construct authorized domain $B_a$ with $D_a$, and unauthorized domain $B_u$ with $D_u$.} $
\STATE $ \textbf{For $epoch=1$ to $Max_{epochs}$ do} $
\STATE $ \text{\qquad Calculate the output of $x_a$, $x_u$ in visual encoder: $f_v^a$, $f_v^u$.} $
\STATE $ \text{\qquad Calculate the augmented feature: $s_v^a$, $s_v^u$} $
\STATE $ \text{\qquad Construct $Prompt_a$ and $Prompt_u$ by multi-scale features from the L-layer visual encoder.} $
\STATE $ \text{\qquad Calculate the output of $Prompt_a$, $Prompt_u$ in test encoder: $f_t^a$, $f_t^u$.} $
\STATE $ \text{\qquad Update $\theta$ by Eq. (12)} $  
\STATE $ \textbf{End For} $
\STATE Return projector parameters $\theta$ and $\phi$.
\end{algorithmic}
\label{alg1}
\end{algorithm*}

\begin{algorithm*}[!t]
\caption{Target-Free IP-CLIP.}\label{alg:alg2}
\begin{algorithmic}[1]
\REQUIRE{The authorized domain $D_a$, visual encoder of CLIP with $L$ layers, text encoder of CLIP, the parameters $\theta$ of IP Projector, the parameters $\phi$ of STAM, augmentation pool $A={\left\{a_i\right\}}_{i=1}^{N_{A}}$, and number of augmentation $n_{aug}<N_{A}$.}
\STATE $ \text{Initialize unauthorized domain $D_u=\emptyset$.} $
\STATE $ \textbf{For $i=1$ to $N_a$ do} $
\STATE $ \textbf{\qquad For $j=1$ to $n_{aug}$ do} $
\STATE $ \text{\qquad\qquad Random select $a_j \in A$} $
\STATE $ \text{\qquad\qquad style augmentation: $x_i \leftarrow a_j(x_i)$} $
\STATE $ \textbf{\qquad End For} $
\STATE $ \text{\qquad Update $D_u = D_u \cup x_i$} $
\STATE $ \textbf{End For} $

\STATE $ \text{Construct authorized domain $B_a$ with $D_a$, and unauthorized domain $B_u$ with $D_u$.} $
\STATE $ \textbf{For $epoch=1$ to $Max_{epochs}$ do} $
\STATE $ \text{\qquad Calculate the output of $x_a$, $x_u$ in in visual encoder: $f_v^a$, $f_v^u$.} $
\STATE $ \text{\qquad Calculate the augmented feature: $s_v^a$, $s_v^u$} $
\STATE $ \text{\qquad Construct $Prompt_a$ and $Prompt_u$ by multi-scale features from the L-layer visual encoder.} $
\STATE $ \text{\qquad Calculate the output of $Prompt_a$, $Prompt_u$ in test encoder: $f_t^a$, $f_t^u$.} $
\STATE $ \text{\qquad Update $\theta$ by Eq. (12).} $  
\STATE $ \textbf{End For} $
\STATE Return projector parameters $\theta$ and $\phi$.
\end{algorithmic}
\label{alg2}
\end{algorithm*}

\begin{table*}[!t]
\renewcommand\arraystretch{1.2}
  \centering
  \resizebox{1\textwidth}{!}{
  \begin{tabular}{c|ccc|ccc}
    \toprule
    Authorized/Unauthorized & Amazon & Dslr & Webcam & $W_{ua} \uparrow$ & $D_u \uparrow$ & $D_a \downarrow$ \\
    \midrule   
    Amazon & 82.1 $\Rightarrow$  79.0 & 79.7 $\Rightarrow$  35.9 & 71.9 $\Rightarrow$ \ 4.7 & 41.37 & 55.50 & 3.10 \\ 
    Dslr   & 65.6 $\Rightarrow$ \ 9.4 & 99.2 $\Rightarrow$  97.7 & 92.2 $\Rightarrow$ \ 0.0 & 70.94 & 74.20 & 1.55 \\ 
    Webcam & 65.6 $\Rightarrow$ \ 3.1 & 93.8 $\Rightarrow$ \ 4.7 & 97.7 $\Rightarrow$  97.7 & 74.02 & 75.80 & 0.00 \\
    \midrule
    Mean   & \multicolumn{3}{c|}{/} & 62.11 & 68.50 & 2.07 \\
    \bottomrule
    \end{tabular}}
    \caption{The accuracy ($\%$) of target-specified NTL [27] on the Office-31 [25]. The vertical/horizontal axis denotes the authorized/unauthorized domain. In each task, the left of '\(\Rightarrow\)' shows the test accuracy of supervised learning CNN on the unauthorized domain, while the right side presents the accuracy of NTL. $W_{ua}$ represents the proposed weighted drop, while $D_u$ and $D_a$ denote the drop rates for the unauthorized and authorized domains, respectively.}
  \label{r_ts_office31_ntl}
\end{table*}

\begin{table*}[!t]
\renewcommand\arraystretch{1.2}
  \centering
  \resizebox{1\textwidth}{!}{
  \begin{tabular}{c|ccc|ccc}
    \toprule
    Authorized/Unauthorized & Amazon & Dslr & Webcam & $W_{ua} \uparrow$ & $D_u \uparrow$ & $D_a \downarrow$ \\
    \midrule   
    Amazon & 82.1 $\Rightarrow$  81.3 & 79.7 $\Rightarrow$ \ 0.0 & 71.9 $\Rightarrow$ \ 0.0 & 60.94 & 75.80 & 0.80 \\ 
    Dslr   & 65.6 $\Rightarrow$ \ 3.1 & 99.2 $\Rightarrow$  98.4 & 92.2 $\Rightarrow$ \ 0.0 & 75.33 & 77.35 & 0.80 \\ 
    Webcam & 65.6 $\Rightarrow$ \ 3.1 & 93.8 $\Rightarrow$ \ 4.7 & 97.7 $\Rightarrow$  97.7 & 74.02 & 75.80 & 0.00 \\
    \midrule
    Mean   & \multicolumn{3}{c|}{/} & 70.09 & 76.32 & 0.53  \\
    \bottomrule
    \end{tabular}}
    \caption{The accuracy ($\%$) of target-specified CUTI-Domain [28] on the Office-31 [25]. The vertical/horizontal axis denotes the authorized/unauthorized domain. In each task, the left of '\(\Rightarrow\)' shows the test accuracy of supervised learning CNN on the unauthorized domain, while the right side presents the accuracy of CUTI-Domain. $W_{ua}$ represents the proposed weighted drop, while $D_u$ and $D_a$ denote the drop rates for the unauthorized and authorized domains, respectively.}
  \label{r_ts_office31_cuti}
\end{table*}

\begin{table*}[!t]
\renewcommand\arraystretch{1.2}
  \centering
  \resizebox{1\textwidth}{!}{
  \begin{tabular}{c|ccc|ccc}
    \toprule
    Authorized/Unauthorized & Amazon & Dslr & Webcam & $W_{ua} \uparrow$ & $D_u \uparrow$ & $D_a \downarrow$ \\
    \midrule   
    Amazon & 79.4 $\Rightarrow$   77.6 & 87.5 $\Rightarrow$ \ 10.0 & 88.8  $\Rightarrow$ \ 17.5 & 56.34 & 74.40 & 1.80 \\ 
    Dslr   & 83.8 $\Rightarrow$ \ 10.0 & 95.7 $\Rightarrow$  94.4 & 98.8 $\Rightarrow$ \ 8.8 & 76.09 & 81.90 & 1.30 \\ 
    Webcam & 80.0 $\Rightarrow$ \ 3.8 & 92.5 $\Rightarrow$ \ 91.3 & 94.4  $\Rightarrow$  91.3 & 32.50 & 38.70 & 3.10 \\
    \midrule
    Mean   & \multicolumn{3}{c|}{/} & 54.98 & 65.00 & 2.07 \\
    \bottomrule
    \end{tabular}}
    \caption{The accuracy ($\%$) of target-specified CLIP-based NTL [27] on the Office-31 [25]. The vertical/horizontal axis denotes the authorized/unauthorized domain. In each task, the left of '\(\Rightarrow\)' shows the test accuracy of supervised learning CLIP on the unauthorized domain, while the right side presents the accuracy of CLIP-based NTL. $W_{ua}$ represents the proposed weighted drop, while $D_u$ and $D_a$ denote the drop rates for the unauthorized and authorized domains, respectively.}
  \label{r_ts_office31_ntl_p}
\end{table*}

\begin{table*}[!t]
\renewcommand\arraystretch{1.2}
  \centering
  \resizebox{1\textwidth}{!}{
  \begin{tabular}{c|ccc|ccc}
    \toprule
    Authorized/Unauthorized & Amazon & Dslr & Webcam & $W_{ua} \uparrow$ & $D_u \uparrow$ & $D_a \downarrow$ \\
    \midrule   
    Amazon & 79.4 $\Rightarrow$  78.8 & 87.5 $\Rightarrow$ \ 6.3 & 88.8  $\Rightarrow$ \ 11.3 & 62.06 & 79.35 & 0.60 \\ 
    Dslr   & 83.8 $\Rightarrow$ \ 5.0 & 95.7 $\Rightarrow$ 95.0 & 98.8 $\Rightarrow$ \ 7.5 & 80.13 & 85.05 & 0.70 \\ 
    Webcam & 80.0 $\Rightarrow$ \ 1.3 & 92.5 $\Rightarrow$ \ 2.5 & 94.4  $\Rightarrow$  91.9 & 75.24 & 84.38 & 2.50 \\
    \midrule
    Mean   & \multicolumn{3}{c|}{/} & 72.48 & 82.93 & 1.27 \\
    \bottomrule
    \end{tabular}}
    \caption{The accuracy ($\%$) of target-specified CLIP-based CUTI-Domain [28] on the Office-31 [25]. The vertical/horizontal axis denotes the authorized/unauthorized domain. In each task, the left of '\(\Rightarrow\)' shows the test accuracy of supervised learning CLIP on the unauthorized domain, while the right side presents the accuracy of CLIP-based CUTI-Domain. $W_{ua}$ represents the proposed weighted drop, while $D_u$ and $D_a$ denote the drop rates for the unauthorized and authorized domains, respectively.}
  \label{r_ts_office31_cuti_p}
\end{table*}

\begin{table*}[!t]
\renewcommand\arraystretch{1.2}
  \centering
  \resizebox{1\textwidth}{!}{
  \begin{tabular}{c|ccc|ccc}
    \toprule
    Authorized/Unauthorized & Amazon & Dslr & Webcam & $W_{ua} \uparrow$ & $D_u \uparrow$ & $D_a \downarrow$ \\
    \midrule   
    Amazon & 79.4 $\Rightarrow$  79.4 & 87.5 $\Rightarrow$ \ 7.5 & 88.8 $\Rightarrow$ \ 8.8 & 63.52 & 80.00 & 0.00 \\ 
    Dslr   & 83.8 $\Rightarrow$ \ 3.8 & 95.7 $\Rightarrow$  95.7 & 98.8 $\Rightarrow$ \ 6.3 & 82.54 & 86.25 & 0.00 \\ 
    Webcam & 80.0 $\Rightarrow$ \ 3.8 & 92.5 $\Rightarrow$ \ 2.5 & 94.4 $\Rightarrow$  94.4 & 78.45 & 83.10 & 0.00 \\
    \midrule
    Mean   & \multicolumn{3}{c|}{/} & 74.84 & 83.12 & 0.00 \\
    \bottomrule
  \end{tabular}}
    \caption{The accuracy ($\%$) of target-specified IP-CLIP on the Office-31 [25]. The vertical/horizontal axis denotes the authorized/unauthorized domain. In each task, the left of '\(\Rightarrow\)' shows the test accuracy of supervised learning CLIP on the unauthorized domain, while the right side presents the accuracy of IP-CLIP. $W_{ua}$ represents the proposed weighted drop, while $D_u$ and $D_a$ denote the drop rates for the unauthorized and authorized domains, respectively.}
  \label{r_ts_office31_detail}
\end{table*}

\begin{table*}[!t]
\renewcommand\arraystretch{1.2}
  \centering
  \resizebox{1\textwidth}{!}{
  \begin{tabular}{c|cccc|ccc}
    \toprule
    Authorized/Unauthorized & Art & Clipart & Product & RealWorld & $W_{ua} \uparrow$ & $D_u \uparrow$ & $D_a \downarrow$ \\
    \midrule   
    Art        &    76.3  $\Rightarrow$ 75.5  & 47.1  $\Rightarrow$ 1.8  & 64.9  $\Rightarrow$ 2.6  & 72.2  $\Rightarrow$ 68.0  & 27.53  & 37.27  & 0.80  \\
    Clipart    &    57.8  $\Rightarrow$ 4.2  & 80.1  $\Rightarrow$ 79.9  & 63.5  $\Rightarrow$ 6.5  & 68.8  $\Rightarrow$ 16.4  & 43.23 & 54.31 & 0.20  \\
    Product    &    56.6  $\Rightarrow$ 6.3  & 45.2  $\Rightarrow$ 3.4  & 92.7  $\Rightarrow$ 92.4  & 72.7  $\Rightarrow$ 29.7  & 41.31  & 45.01 & 0.30 \\
    RealWorld  &    63.8  $\Rightarrow$ 26.6  & 49.2  $\Rightarrow$ 6.5  & 75.5  $\Rightarrow$ 64.3  & 84.4  $\Rightarrow$ 82.0  & 22.93 & 30.37 & 2.40 \\  
    \hline
    Mean       & \multicolumn{4}{c|}{/} &33.75 & 41.74 & 0.43 \\  
    \bottomrule
  \end{tabular}}
      \caption{The accuracy ($\%$) of target-specified NTL [27] on the Office-Home-65~[26]. The vertical/horizontal axis denotes the authorized/unauthorized domain. In each task, the left of '\(\Rightarrow\)' shows the test accuracy of supervised learning CNN on the unauthorized domain, while the right side presents the accuracy of NTL. $W_{ua}$ represents the proposed weighted drop, while $D_u$ and $D_a$ denote the drop rates for the unauthorized and authorized domains, respectively.}
  \label{r_ts_home_ntl}
\end{table*}

\begin{table*}[!t]
\renewcommand\arraystretch{1.2}
  \centering
  \resizebox{1\textwidth}{!}{
  \begin{tabular}{c|cccc|ccc}
    \toprule
    Authorized/Unauthorized & Art & Clipart & Product & RealWorld & $W_{ua} \uparrow$ & $D_u \uparrow$ & $D_a \downarrow$ \\
    \midrule   
    Art        &    76.3  $\Rightarrow$ 76.0  & 47.1  $\Rightarrow$ 4.4  & 64.9  $\Rightarrow$ 9.4  & 72.2  $\Rightarrow$ 28.9  & 35.62 & 47.16 & 0.30 \\
    Clipart    &    57.8  $\Rightarrow$ 4.4  & 80.1  $\Rightarrow$ 79.9  & 63.5  $\Rightarrow$ 5.5  & 68.8  $\Rightarrow$ 8.1  & 45.67 & 57.35 & 0.20  \\
    Product    &    56.6  $\Rightarrow$ 9.1  & 45.2  $\Rightarrow$ 4.2  & 92.7  $\Rightarrow$ 92.2  & 72.7  $\Rightarrow$ 23.7  & 41.78 & 45.82  & 0.50  \\
    RealWorld  &    63.8  $\Rightarrow$ 31.3  & 49.2  $\Rightarrow$ 8.6  & 75.5  $\Rightarrow$ 19.8  & 84.4  $\Rightarrow$ 84.1  & 35.87 & 42.95 & 0.30  \\
    \hline
    Mean       & \multicolumn{4}{c|}{/} &39.73  & 48.32 & 0.33 \\  
    \bottomrule
  \end{tabular}}
      \caption{The accuracy ($\%$) of target-specified CUTI-Domain [28] on the Office-Home-65~[26]. The vertical/horizontal axis denotes the authorized/unauthorized domain. In each task, the left of '\(\Rightarrow\)' shows the test accuracy of supervised learning CNN on the unauthorized domain, while the right side presents the accuracy of CUTI-Domain. $W_{ua}$ represents the proposed weighted drop, while $D_u$ and $D_a$ denote the drop rates for the unauthorized and authorized domains, respectively.}
  \label{r_ts_home_cuti}
\end{table*}

\begin{table*}[!t]
\renewcommand\arraystretch{1.2}
  \centering
  \resizebox{1\textwidth}{!}{
  \begin{tabular}{c|cccc|ccc}
    \toprule
    Authorized/Unauthorized & Art & Clipart & Product & RealWorld & $W_{ua} \uparrow$ & $D_u \uparrow$ & $D_a \downarrow$ \\
    \midrule   
    Art        & 85.5 $\Rightarrow$ 85.4 & 68.0 $\Rightarrow$  23.8 & 89.8 $\Rightarrow$ 86.5 & 88.5 $\Rightarrow$ 88.5 & 13.44 & 15.83 & 0.10 \\ 
    Clipart    & 81.0 $\Rightarrow$ 18.3 & 75.0 $\Rightarrow$  74.7 & 90.8 $\Rightarrow$ \ 15.0 & 89.5 $\Rightarrow$ 31.0 & 48.83 & 65.67 & 0.30  \\ 
    Product    & 78.8 $\Rightarrow$ 11.0 & 73.3 $\Rightarrow$ \ 13.8 & 92.8 $\Rightarrow$  92.8 & 87.5 $\Rightarrow$ 85.8 & 39.90 & 43.00 & 0.00 \\
    RealWorld  & 83.0 $\Rightarrow$ 80.8 & 71.3 $\Rightarrow$ \ 7.8 & 90.8 $\Rightarrow$  52.5 & 90.0 $\Rightarrow$ 88.1 & 28.87 & 34.67 & 1.90 \\
    \hline
    Mean       & \multicolumn{4}{c|}{/} &32.76 & 39.79 & 0.57 \\  
    \bottomrule
  \end{tabular}}
      \caption{The accuracy ($\%$) of target-specified CLIP-based NTL [27] on the Office-Home-65~[26]. The vertical/horizontal axis denotes the authorized/unauthorized domain. In each task, the left of '\(\Rightarrow\)' shows the test accuracy of supervised learning CLIP on the unauthorized domain, while the right side presents the accuracy of CLIP-based NTL. $W_{ua}$ represents the proposed weighted drop, while $D_u$ and $D_a$ denote the drop rates for the unauthorized and authorized domains, respectively.}
  \label{r_ts_home_ntl_p}
\end{table*}

\begin{table*}[!t]
\renewcommand\arraystretch{1.2}
  \centering
  \resizebox{1\textwidth}{!}{
  \begin{tabular}{c|cccc|ccc}
    \toprule
    Authorized/Unauthorized & Art & Clipart & Product & RealWorld & $W_{ua} \uparrow$ & $D_u \uparrow$ & $D_a \downarrow$ \\
    \midrule   
    Art        & 85.5 $\Rightarrow$ 82.5 & 68.0 $\Rightarrow$  16.8 & 89.8 $\Rightarrow$ 21.3 & 88.5 $\Rightarrow$ 48.0 & 41.58 & 53.40 & 3.00 \\ 
    Clipart    & 81.0 $\Rightarrow$ 15.8 & 75.0 $\Rightarrow$  75.0 & 90.8 $\Rightarrow$ \ 9.0 & 89.5 $\Rightarrow$ 19.3 & 53.37 & 72.40 & 0.63  \\ 
    Product    & 78.8 $\Rightarrow$ 16.3 & 73.3 $\Rightarrow$ \ 10.8 & 92.8 $\Rightarrow$  92.4 & 87.5 $\Rightarrow$ 27.0 & 56.82 & 61.83 & 0.37 \\
    RealWorld  & 83.0 $\Rightarrow$ 30.3 & 71.3 $\Rightarrow$ \ 10.0 & 90.8 $\Rightarrow$  32.8 & 90.0 $\Rightarrow$ 88.5 & 49.41 & 57.33 & 1.50 \\
    \hline
    Mean       & \multicolumn{4}{c|}{/} & 50.29 & 61.24 & 1.38 \\  
    \bottomrule
  \end{tabular}}
      \caption{The accuracy ($\%$) of target-specified CLIP-based CUTI-Domain [28] on the Office-Home-65~[26]. The vertical/horizontal axis denotes the authorized/unauthorized domain. In each task, the left of '\(\Rightarrow\)' shows the test accuracy of supervised learning CLIP on the unauthorized domain, while the right side presents the accuracy of CLIP-based CUTI-Domain. $W_{ua}$ represents the proposed weighted drop, while $D_u$ and $D_a$ denote the drop rates for the unauthorized and authorized domains, respectively.}
  \label{r_ts_home_cuti_p}
\end{table*}

\begin{table*}[!t]
\renewcommand\arraystretch{1.2}
  \centering
  \resizebox{1\textwidth}{!}{
  \begin{tabular}{c|cccc|ccc}
    \toprule
    Authorized/Unauthorized & Art & Clipart & Product & RealWorld & $W_{ua} \uparrow$ & $D_u \uparrow$ & $D_a \downarrow$ \\
    \midrule   
    Art        & 85.5 $\Rightarrow$ 85.2 & 68.0 $\Rightarrow$  12.8 & 89.8 $\Rightarrow$  15.0 & 88.5 $\Rightarrow$ 34.5 & 52.00 & 61.33 & 0.30 \\ 
    Clipart    & 81.0 $\Rightarrow$ 11.8 & 75.0 $\Rightarrow$  74.9 & 90.8 $\Rightarrow$ \ 5.3 & 89.5 $\Rightarrow$ 17.8 & 56.45 & 75.47 & 0.10 \\ 
    Product    & 78.8 $\Rightarrow$ 14.0 & 73.3 $\Rightarrow$ \ 8.5 & 92.8 $\Rightarrow$  92.5 & 87.5 $\Rightarrow$ 25.8 & 58.71 & 63.77 & 0.30 \\
    RealWorld  & 83.0 $\Rightarrow$ 30.3 & 71.3 $\Rightarrow$ \ 7.5 & 90.8 $\Rightarrow$  29.3 & 90.0 $\Rightarrow$ 89.9 & 53.25 & 59.33 & 0.10 \\
    \hline
    Mean       & \multicolumn{4}{c|}{/} & 55.10 & 64.98 & 0.20 \\  
    \bottomrule
  \end{tabular}}
      \caption{The accuracy ($\%$) of target-specified IP-CLIP on the Office-Home-65~[26]. The vertical/horizontal axis denotes the authorized/unauthorized domain. In each task, the left of '\(\Rightarrow\)' shows the test accuracy of supervised learning CLIP on the unauthorized domain, while the right side presents the accuracy of IP-CLIP. $W_{ua}$ represents the proposed weighted drop, while $D_u$ and $D_a$ denote the drop rates for the unauthorized and authorized domains, respectively.}
  \label{r_ts_home_detail}
\end{table*}

\begin{table*}[!t]
\renewcommand\arraystretch{1.2}
  \centering
  \resizebox{1\textwidth}{!}{
  \begin{tabular}{c|cccc|cccc}
    \toprule
    Authorized/Unauthorized & Art & Clipart & Product & Real & $W_{ua} \uparrow$ & $D_u \uparrow$ & $D_a \downarrow$ \\
    \midrule   
        Clipart   &  76.4  $\Rightarrow$ 74.3  & 49.5  $\Rightarrow$ 12.7  & 62.5  $\Rightarrow$ 27.0  & 47.0  $\Rightarrow$ 9.5  & 25.63 & 36.60 & 2.10 \\
        Painting  &   48.0  $\Rightarrow$ 5.8  & 61.8  $\Rightarrow$ 61.3  & 60.4  $\Rightarrow$ 38.3  & 42.3  $\Rightarrow$ 9.5  & 19.53 & 32.37 & 0.50 \\
        Real      &  52.3  $\Rightarrow$ 16.6  & 57.9  $\Rightarrow$ 27.8  & 85.6  $\Rightarrow$ 84.4  & 46.9  $\Rightarrow$ 5.1  & 29.26 & 35.87 & 1.20 \\
        Sketch    &  52.5  $\Rightarrow$ 13.7  & 46.9  $\Rightarrow$ 5.9  & 62.8  $\Rightarrow$ 5.3  & 66.6  $\Rightarrow$ 65.6  & 29.37 & 45.77 & 1.00 \\
    \hline
    Mean      & \multicolumn{4}{c|}{/} & 25.95 & 37.65 & 1.27 \\  
    \bottomrule
  \end{tabular}}
    \caption{The accuracy ($\%$) of target-specified NTL [27] on the Mini-DomainNet [31]. The vertical/horizontal axis denotes the authorized/unauthorized domain. In each task, the left of '\(\Rightarrow\)' shows the test accuracy of supervised learning CNN on the unauthorized domain, while the right side presents the accuracy of NTL. $W_{ua}$ represents the proposed weighted drop, while $D_u$ and $D_a$ denote the drop rates for the unauthorized and authorized domains, respectively.}
  \label{r_ts_mini_ntl}
\end{table*}

\begin{table*}[!t]
\renewcommand\arraystretch{1.2}
  \centering
  \resizebox{1\textwidth}{!}{
  \begin{tabular}{c|cccc|cccc}
    \toprule
    Authorized/Unauthorized & Art & Clipart & Product & Real & $W_{ua} \uparrow$ & $D_u \uparrow$ & $D_a \downarrow$ \\
    \midrule   
     Clipart   &   76.4  $\Rightarrow$ 75.6  & 49.5  $\Rightarrow$ 8.4  & 62.5  $\Rightarrow$ 19.9  & 47.0  $\Rightarrow$ 8.1  & 30.29 & 40.87 & 0.80 \\
     Painting  &   48.0  $\Rightarrow$ 7.9  & 61.8  $\Rightarrow$ 61.1  & 60.4  $\Rightarrow$ 36.8  & 42.3  $\Rightarrow$ 6.3  & 19.88 & 33.23 & 0.70 \\
     Real      &   52.3  $\Rightarrow$ 14.0  & 57.9  $\Rightarrow$ 22.0  & 85.6  $\Rightarrow$ 84.5  & 46.9  $\Rightarrow$ 5.9  & 31.52 & 38.40 & 1.10 \\
     Sketch    &   52.5  $\Rightarrow$ 11.2  & 46.9  $\Rightarrow$ 5.4  & 62.8  $\Rightarrow$ 4.9  & 66.7  $\Rightarrow$ 65.7  & 30.18 & 46.90 & 0.96 \\ 
    \hline
    Mean      & \multicolumn{4}{c|}{/} & 27.97 & 39.85 & 0.87 \\  
    \bottomrule
  \end{tabular}}
    \caption{The accuracy ($\%$) of target-specified CUTI-Domain [28] on the Mini-DomainNet [31]. The vertical/horizontal axis denotes the authorized/unauthorized domain. In each task, the left of '\(\Rightarrow\)' shows the test accuracy of supervised learning CNN on the unauthorized domain, while the right side presents the accuracy of CUTI-Domain. $W_{ua}$ represents the proposed weighted drop, while $D_u$ and $D_a$ denote the drop rates for the unauthorized and authorized domains, respectively.}
  \label{r_ts_mini_cuti}
\end{table*}

\begin{table*}[!t]
\renewcommand\arraystretch{1.2}
  \centering
  \resizebox{1\textwidth}{!}{
  \begin{tabular}{c|cccc|cccc}
    \toprule
    Authorized/Unauthorized & Art & Clipart & Product & Real & $W_{ua} \uparrow$ & $D_u \uparrow$ & $D_a \downarrow$ \\
    \midrule   
    Clipart   & 85.1 $\Rightarrow$ 84.5  & 79.8  $\Rightarrow$ 33.3 & 89.8  $\Rightarrow$ 34.0 & 78.7  $\Rightarrow$ 42.1 & 38.62 & 46.30 & 0.60 \\
    Painting  & 83.8 $\Rightarrow$ 11.9  & 81.4  $\Rightarrow$ 79.8 & 89.1  $\Rightarrow$ 60.5 & 78.4  $\Rightarrow$ 17.5 & 41.66 & 53.80 & 1.60 \\
    Real      & 84.6 $\Rightarrow$ 26.4  & 80.5  $\Rightarrow$ 30.8 & 90.6  $\Rightarrow$ 89.8 & 80.0  $\Rightarrow$ 10.8 & 52.29 & 59.03 & 0.80 \\
    Sketch    & 84.3 $\Rightarrow$ 83.0  & 79.1  $\Rightarrow$ 32.7 & 90.3  $\Rightarrow$ 9.7  & 80.7  $\Rightarrow$ 80.1 & 33.78 & 42.77 & 0.60 \\
    \hline
    Mean      & \multicolumn{4}{c|}{/} & 41.59 & 50.48 & 1.00  \\  
    \bottomrule
  \end{tabular}}
    \caption{The accuracy ($\%$) of target-specified CLIP-based NTL [27] on the Mini-DomainNet [31]. The vertical/horizontal axis denotes the authorized/unauthorized domain. In each task, the left of '\(\Rightarrow\)' shows the test accuracy of supervised learning CLIP on the unauthorized domain, while the right side presents the accuracy of CLIP-based NTL. $W_{ua}$ represents the proposed weighted drop, while $D_u$ and $D_a$ denote the drop rates for the unauthorized and authorized domains, respectively.}
  \label{r_ts_mini_ntl_p}
\end{table*}

\begin{table*}[!t]
\renewcommand\arraystretch{1.2}
  \centering
  \resizebox{1\textwidth}{!}{
  \begin{tabular}{c|cccc|cccc}
    \toprule
    Authorized/Unauthorized & Art & Clipart & Product & Real & $W_{ua} \uparrow$ & $D_u \uparrow$ & $D_a \downarrow$ \\
    \midrule   
        Clipart   & 85.1  $\Rightarrow$ 84.9  & 79.8  $\Rightarrow$ 22.7  & 89.8  $\Rightarrow$ 23.7  & 78.7  $\Rightarrow$ 23.7  & 50.26 & 59.40 & 0.20 \\
        Painting  & 83.8  $\Rightarrow$ 17.9  & 81.4  $\Rightarrow$ 76.1  & 89.1  $\Rightarrow$ 15.6  & 78.4  $\Rightarrow$ 17.1  & 46.88 & 66.90 & 5.30 \\
        Real      & 84.6  $\Rightarrow$ 24.9  & 80.5  $\Rightarrow$ 23.8  & 90.6  $\Rightarrow$ 89.5  & 80.0  $\Rightarrow$ 9.5  & 54.77 & 62.30 & 1.10 \\
        Sketch    & 84.3  $\Rightarrow$ 24.0  & 79.1  $\Rightarrow$ 25.2  & 90.3  $\Rightarrow$ 10.8  & 80.7  $\Rightarrow$ 80.0  & 51.09 & 64.57 & 0.70 \\ 
    \hline
    Mean      & \multicolumn{4}{c|}{/} & 50.75 & 63.29 & 2.20 \\  
    \bottomrule
  \end{tabular}}
    \caption{The accuracy ($\%$) of target-specified CLIP-based CUTI-Domain [28] on the Mini-DomainNet [31]. The vertical/horizontal axis denotes the authorized/unauthorized domain. In each task, the left of '\(\Rightarrow\)' shows the test accuracy of supervised learning CLIP on the unauthorized domain, while the right side presents the accuracy of CLIP-based CUTI-Domain. $W_{ua}$ represents the proposed weighted drop, while $D_u$ and $D_a$ denote the drop rates for the unauthorized and authorized domains, respectively.}
  \label{r_ts_mini_cuti_p}
\end{table*}

\begin{table*}[!t]
\renewcommand\arraystretch{1.2}
  \centering
  \resizebox{1\textwidth}{!}{
  \begin{tabular}{c|cccc|cccc}
    \toprule
    Authorized/Unauthorized & Art & Clipart & Product & Real & $W_{ua} \uparrow$ & $D_u \uparrow$ & $D_a \downarrow$ \\
    \midrule   
    Clipart   & 85.1 $\Rightarrow$  84.8 & 79.8 $\Rightarrow$ 18.1 & 89.8 $\Rightarrow$  24.8 & 78.7 $\Rightarrow$  22.4 & 51.47 & 61.00 & 0.30 \\ 
    Painting  & 83.8 $\Rightarrow$ \ 9.2 & 81.4 $\Rightarrow$ 80.9 & 89.1 $\Rightarrow$  26.5 & 78.4 $\Rightarrow$  14.4 & 53.85 & 67.07 & 0.50 \\ 
    Real      & 84.6 $\Rightarrow$  21.9 & 80.5 $\Rightarrow$ 21.4 & 90.6 $\Rightarrow$  90.4 & 80.0 $\Rightarrow$ \ 6.0 & 58.82 & 65.27 & 0.20 \\
    Sketch    & 84.3 $\Rightarrow$  23.2 & 79.1 $\Rightarrow$ 15.6 & 90.3 $\Rightarrow$ \ 9.2 & 80.7 $\Rightarrow$  80.2 & 54.59 & 68.57 & 0.50 \\
    \hline
    Mean      & \multicolumn{4}{c|}{/} & 54.68& 65.48 & 0.33  \\  
    \bottomrule
  \end{tabular}}
    \caption{The accuracy ($\%$) of target-specified IP-CLIP on the Mini-DomainNet [31]. The vertical/horizontal axis denotes the authorized/unauthorized domain. In each task, the left of '\(\Rightarrow\)' shows the test accuracy of supervised learning CLIP on the unauthorized domain, while the right side presents the accuracy of IP-CLIP. $W_{ua}$ represents the proposed weighted drop, while $D_u$ and $D_a$ denote the drop rates for the unauthorized and authorized domains, respectively.}
  \label{r_ts_mini_detail}
\end{table*}

\begin{table*}[!t]
\renewcommand\arraystretch{1.2}
  \centering
  \resizebox{1\textwidth}{!}{
  \begin{tabular}{c|ccc|ccc}
    \toprule
    Authorized/Test & Amazon & Dslr & Webcam & $W_{ua} \uparrow$ & $D_u \uparrow$ & $D_a \downarrow$ \\
    \midrule   
     Amazon &   82.1  $\Rightarrow$ 75.0  & 79.7  $\Rightarrow$ 64.1  & 71.9  $\Rightarrow$ 71.9  & 0.56  & 7.80  & 7.05\\ 
     Dslr   &   65.6  $\Rightarrow$ 48.4  & 99.2  $\Rightarrow$ 96.9  & 92.2  $\Rightarrow$ 90.6  & 6.88  & 9.40  &  2.30\\ 
     Webcam &    65.6  $\Rightarrow$ 57.8  & 93.8  $\Rightarrow$ 84.4  & 97.7  $\Rightarrow$ 92.2  & 2.90 & 8.60  & 5.45   \\ 
    \midrule
    Mean   & \multicolumn{3}{c|}{/} & 3.45 & 8.60 & 4.93 \\
    \bottomrule
    \end{tabular}}
    \caption{The accuracy ($\%$) of target-free NTL [27] on the Office-31 [25]. The vertical/horizontal axis denotes the authorized/test domain. In each task, the left of '\(\Rightarrow\)' shows the test accuracy of supervised learning CNN on the test domain, while the right side presents the accuracy of NTL. $W_{ua}$ represents the proposed weighted drop, while $D_u$ and $D_a$ denote the drop rates for the unauthorized and authorized domains, respectively.
}
  \label{r_tf_office31_ntl}
\end{table*}

\begin{table*}[!t]
\renewcommand\arraystretch{1.2}
  \centering
  \resizebox{1\textwidth}{!}{
  \begin{tabular}{c|ccc|ccc}
    \toprule
    Authorized/Test & Amazon & Dslr & Webcam & $W_{ua} \uparrow$ & $D_u \uparrow$ & $D_a \downarrow$ \\
    \midrule   
        Amazon & 82.1  $\Rightarrow$ 75.0  & 79.7  $\Rightarrow$ 64.1  & 71.9  $\Rightarrow$ 60.9  & 4.69   & 13.30  & 7.05 \\
        Dslr & 65.6  $\Rightarrow$ 46.9  & 99.2  $\Rightarrow$ 96.9  & 92.2  $\Rightarrow$ 92.2  & 6.83   & 9.35  & 2.30 \\
        Webcam & 65.6  $\Rightarrow$ 59.4  & 93.8  $\Rightarrow$ 89.1  & 97.7  $\Rightarrow$ 95.3  & 2.95   & 5.45  & 2.35 \\
    \midrule
    Mean   & \multicolumn{3}{c|}{/} & 4.82 & 9.37 & 3.90 \\
    \bottomrule
    \end{tabular}}
    \caption{The accuracy ($\%$) of target-free CUTI-Domain [28] on the Office-31 [25]. The vertical/horizontal axis denotes the authorized/test domain. In each task, the left of '\(\Rightarrow\)' shows the test accuracy of supervised learning CNN on the test domain, while the right side presents the accuracy of CUTI-Domain. $W_{ua}$ represents the proposed weighted drop, while $D_u$ and $D_a$ denote the drop rates for the unauthorized and authorized domains, respectively.}
  \label{r_tf_office31_cuti}
\end{table*}

\begin{table*}[!t]
\renewcommand\arraystretch{1.2}
  \centering
  \resizebox{1\textwidth}{!}{
  \begin{tabular}{c|ccc|ccc}
    \toprule
    Authorized/Test & Amazon & Dslr & Webcam & $W_{ua} \uparrow$ & $D_u \uparrow$ & $D_a \downarrow$ \\
    \midrule   
        Amazon & 79.4  $\Rightarrow$ 77.5  & 87.5  $\Rightarrow$ 62.5  & 88.8  $\Rightarrow$ 79.3  & 11.90   & 17.25  & 1.90 \\
        Dslr & 83.8  $\Rightarrow$ 27.0  & 95.7  $\Rightarrow$ 91.8  & 98.8  $\Rightarrow$ 67.8  & 36.72   & 43.90  & 3.90 \\ 
        Webcam & 80.0  $\Rightarrow$ 23.8  & 92.5  $\Rightarrow$ 46.8  & 94.4  $\Rightarrow$ 92.8  & 45.80   & 50.95  & 1.60 \\
    \midrule
    Mean   & \multicolumn{3}{c|}{/} & 31.47 & 37.37 & 2.47 \\
    \bottomrule
    \end{tabular}}
    \caption{The accuracy ($\%$) of target-free CLIP-based NTL [27] on the Office-31 [25]. The vertical/horizontal axis denotes the authorized/test domain. In each task, the left of '\(\Rightarrow\)' shows the test accuracy of supervised learning CLIP on the test domain, while the right side presents the accuracy of CLIP-based NTL. $W_{ua}$ represents the proposed weighted drop, while $D_u$ and $D_a$ denote the drop rates for the unauthorized and authorized domains, respectively.}
  \label{r_tf_office31_ntl_p}
\end{table*}

\begin{table*}[!t]
\renewcommand\arraystretch{1.2}
  \centering
  \resizebox{1\textwidth}{!}{
  \begin{tabular}{c|ccc|ccc}
    \toprule
    Authorized/Test & Amazon & Dslr & Webcam & $W_{ua} \uparrow$ & $D_u \uparrow$ & $D_a \downarrow$ \\
    \midrule   
        Amazon & 79.4  $\Rightarrow$  76.3  & 87.5  $\Rightarrow$  34.5  & 88.8  $\Rightarrow$  68.5  & 25.60   & 36.65  & 3.10 \\ 
        Dslr & 83.8  $\Rightarrow$  15.0  & 95.7  $\Rightarrow$  93.8  & 98.8  $\Rightarrow$  81.0  & 38.83   & 43.30  & 1.90 \\ 
        Webcam & 80.0  $\Rightarrow$  77.8  & 92.5  $\Rightarrow$  27.5  & 94.4  $\Rightarrow$  93.8  & 30.95   & 33.60  & 0.60 \\ 
    \midrule
        Mean & \multicolumn{3}{c|}{/} & 31.80   & 37.85  & 1.87 \\ 
    \bottomrule
    \end{tabular}}
    \caption{The accuracy ($\%$) of target-free CLIP-based CUTI-Domain [28] on the Office-31 [25]. The vertical/horizontal axis denotes the authorized/test domain. In each task, the left of '\(\Rightarrow\)' shows the test accuracy of supervised learning CLIP on the test domain, while the right side presents the accuracy of CLIP-based CUTI-Domain. $W_{ua}$ represents the proposed weighted drop, while $D_u$ and $D_a$ denote the drop rates for the unauthorized and authorized domains, respectively.}
  \label{r_tf_office31_cuti_p}
\end{table*}

\begin{table*}[!t]
\renewcommand\arraystretch{1.2}
  \centering
  \resizebox{1\textwidth}{!}{
  \begin{tabular}{c|ccc|ccc}
    \toprule
    Authorized/Test & Amazon & Dslr & Webcam & $W_{ua} \uparrow$ & $D_u \uparrow$ & $D_a \downarrow$ \\
    \midrule   
        Amazon & 79.4  $\Rightarrow$  79.0  & 87.5  $\Rightarrow$  9.8  & 88.8  $\Rightarrow$  38.3  & 50.32   & 64.10  & 0.40  \\
        Dslr & 83.8  $\Rightarrow$  23.3  & 95.7  $\Rightarrow$  95.3  & 98.8  $\Rightarrow$  64.3  & 44.89   & 47.50  & 0.40 \\
        Webcam &  80.0  $\Rightarrow$  17.8  & 92.5  $\Rightarrow$  10.0  & 94.4  $\Rightarrow$  92.5  & 65.17   & 72.35  & 1.90 \\
    \midrule
        Mean & \multicolumn{3}{c|}{/} & 53.46   & 61.32  & 0.90  \\
    \bottomrule
  \end{tabular}}
    \caption{The accuracy ($\%$) of target-specified IP-CLIP on the Office-31 [25]. The vertical/horizontal axis denotes the authorized/test domain. In each task, the left of '\(\Rightarrow\)' shows the test accuracy of supervised learning CLIP on the test domain, while the right side presents the accuracy of IP-CLIP. $W_{ua}$ represents the proposed weighted drop, while $D_u$ and $D_a$ denote the drop rates for the unauthorized and authorized domains, respectively.}
  \label{r_tf_office31_detail}
\end{table*}

\begin{table*}[!t]
\renewcommand\arraystretch{1.2}
  \centering
  \resizebox{1\textwidth}{!}{
  \begin{tabular}{c|cccc|ccc}
    \toprule
    Authorized/Test & Art & Clipart & Product & RealWorld & $W_{ua} \uparrow$ & $D_u \uparrow$ & $D_a \downarrow$ \\
    \midrule   
        Art & 76.3  $\Rightarrow$  74.5  & 47.1  $\Rightarrow$  43.5  & 64.9  $\Rightarrow$  63.3  & 72.2  $\Rightarrow$  71.6  & 0.10   & 1.93  & 1.80\\
        Clipart & 57.8  $\Rightarrow$  55.7  & 80.1  $\Rightarrow$  79.7  & 63.5  $\Rightarrow$  61.5  & 68.8  $\Rightarrow$  68.8  & 0.75   & 1.34  & 0.40 \\
        Product & 56.6  $\Rightarrow$  51.0  & 45.2  $\Rightarrow$  37.0  & 92.7  $\Rightarrow$  90.1  & 72.7  $\Rightarrow$  68.2  & 3.13   & 6.08  & 2.60 \\
        RealWorld & 63.8  $\Rightarrow$  62.5  & 49.2  $\Rightarrow$  43.8  & 75.5  $\Rightarrow$  73.7  & 84.4  $\Rightarrow$  84.4  & 2.39   & 2.83  & 0.00 \\
    \hline
    Mean       & \multicolumn{4}{c|}{/} & 1.59   & 3.05  & 1.20  \\  
    \bottomrule
  \end{tabular}}
      \caption{The accuracy ($\%$) of target-free NTL [27] on the Office-Home-65~[26]. The vertical/horizontal axis denotes the authorized/test domain. In each task, the left of '\(\Rightarrow\)' shows the test accuracy of supervised learning CNN on the test domain, while the right side presents the accuracy of NTL. $W_{ua}$ represents the proposed weighted drop, while $D_u$ and $D_a$ denote the drop rates for the unauthorized and authorized domains, respectively.}
  \label{r_tf_home_ntl}
\end{table*}

\begin{table*}[!t]
\renewcommand\arraystretch{1.2}
  \centering
  \resizebox{1\textwidth}{!}{
  \begin{tabular}{c|cccc|ccc}
    \toprule
    Authorized/Test & Art & Clipart & Product & RealWorld & $W_{ua} \uparrow$ & $D_u \uparrow$ & $D_a \downarrow$ \\
    \midrule   
        Art & 76.3  $\Rightarrow$  69.5  & 47.1  $\Rightarrow$  39.3  & 64.9  $\Rightarrow$  56.8  & 72.2  $\Rightarrow$  68.5  & -0.19   & 6.53  &  6.80  \\
        Clipart & 57.8  $\Rightarrow$  44.5  & 80.1  $\Rightarrow$  73.7  & 63.5  $\Rightarrow$  59.1  & 68.8  $\Rightarrow$  61.7  & 1.36   &  8.24  & 6.40 \\
        Product & 56.6  $\Rightarrow$  42.2  & 45.2  $\Rightarrow$  31.3  & 92.7  $\Rightarrow$  84.6  & 72.7  $\Rightarrow$  61.7  & 4.21   & 13.08  & 8.10  \\
        RealWorld & 63.8  $\Rightarrow$  53.1  & 49.2  $\Rightarrow$  40.4  & 75.5  $\Rightarrow$  68.5  & 84.4  $\Rightarrow$  80.2  & 3.72   &8.83 & 4.20\\  
    \hline
    Mean       & \multicolumn{4}{c|}{/} & 2.28 & 9.17 & 6.38 \\  
    \bottomrule
  \end{tabular}}
      \caption{The accuracy ($\%$) of target-free CUTI-Domain [28] on the Office-Home-65~[26]. The vertical/horizontal axis denotes the authorized/test domain. In each task, the left of '\(\Rightarrow\)' shows the test accuracy of supervised learning CNN on the test domain, while the right side presents the accuracy of CUTI-Domain. $W_{ua}$ represents the proposed weighted drop, while $D_u$ and $D_a$ denote the drop rates for the unauthorized and authorized domains, respectively.}
  \label{r_tf_home_cuti}
\end{table*}

\begin{table*}[!t]
\renewcommand\arraystretch{1.2}
  \centering
  \resizebox{1\textwidth}{!}{
  \begin{tabular}{c|cccc|ccc}
    \toprule
    Authorized/Test & Art & Clipart & Product & RealWorld & $W_{ua} \uparrow$ & $D_u \uparrow$ & $D_a \downarrow$ \\
    \midrule   
        Art & 85.5  $\Rightarrow$  81.8  & 68.0  $\Rightarrow$  64.5  & 89.8  $\Rightarrow$  88.3  & 88.5  $\Rightarrow$  85.0  & -0.71   & 2.83  & 3.70  \\
        Clipart & 81.0  $\Rightarrow$  78.8  & 75.0  $\Rightarrow$  74.5  & 90.8  $\Rightarrow$  91.8  & 89.5  $\Rightarrow$  88.0  & 0.30   & 0.90  & 0.50   \\
        Product & 78.8  $\Rightarrow$  60.0  & 73.3  $\Rightarrow$  35.0  & 92.8  $\Rightarrow$  89.5  & 87.5  $\Rightarrow$  87.5  & 14.08   & 19.03  & 3.30  \\
        RealWorld & 83.0  $\Rightarrow$  71.3  & 71.3  $\Rightarrow$  31.8  & 90.8  $\Rightarrow$  89.0  & 90.0  $\Rightarrow$  87.3  & 13.07   & 17.67  & 2.70  \\
    \hline
    Mean       & \multicolumn{4}{c|}{/} &6.68   & 10.11  & 2.55  \\  
    \bottomrule
  \end{tabular}}
      \caption{The accuracy ($\%$) of target-free CLIP-based NTL [27] on the Office-Home-65~[26]. The vertical/horizontal axis denotes the authorized/test domain. In each task, the left of '\(\Rightarrow\)' shows the test accuracy of supervised learning CLIP on the test domain, while the right side presents the accuracy of CLIP-based NTL. $W_{ua}$ represents the proposed weighted drop, while $D_u$ and $D_a$ denote the drop rates for the unauthorized and authorized domains, respectively.}
  \label{r_tf_home_ntl_p}
\end{table*}

\begin{table*}[!t]
\renewcommand\arraystretch{1.2}
  \centering
  \resizebox{1\textwidth}{!}{
  \begin{tabular}{c|cccc|ccc}
    \toprule
    Authorized/Test & Art & Clipart & Product & RealWorld & $W_{ua} \uparrow$ & $D_u \uparrow$ & $D_a \downarrow$ \\
    \midrule   
        Art & 85.5  $\Rightarrow$  81.3  & 68.0  $\Rightarrow$  62.5  & 89.8  $\Rightarrow$  89.8  & 88.5  $\Rightarrow$  83.8  & -0.65   & 3.40  & 4.20  \\
        Clipart & 81.0  $\Rightarrow$  70.8  & 75.0  $\Rightarrow$  73.8  & 90.8  $\Rightarrow$  87.3  & 89.5  $\Rightarrow$  78.5  & 5.19   & 8.23  & 1.20 \\
        Product & 78.8  $\Rightarrow$  67.3  & 73.3  $\Rightarrow$  49.8  & 92.8  $\Rightarrow$  88.5  & 87.5  $\Rightarrow$  67.0  & 12.57   & 18.50   & 4.30 \\
        RealWorld & 83.0  $\Rightarrow$  79.5  & 71.3  $\Rightarrow$  62.8  & 90.8  $\Rightarrow$  86.3  & 90.0  $\Rightarrow$  88.8  & 3.82   & 5.50  & 1.20 \\
    \hline
    Mean       & \multicolumn{4}{c|}{/} & 5.23   & 8.91  &2.73  \\  
    \bottomrule
  \end{tabular}}
      \caption{The accuracy ($\%$) of target-free CLIP-based CUTI-Domain [28] on the Office-Home-65~[26]. TThe vertical/horizontal axis denotes the authorized/test domain. In each task, the left of '\(\Rightarrow\)' shows the test accuracy of supervised learning CLIP on the test domain, while the right side presents the accuracy of CLIP-based CUTI-Domain. $W_{ua}$ represents the proposed weighted drop, while $D_u$ and $D_a$ denote the drop rates for the unauthorized and authorized domains, respectively.}
  \label{r_tf_home_cuti_p}
\end{table*}

\begin{table*}[!t]
\renewcommand\arraystretch{1.2}
  \centering
  \resizebox{1\textwidth}{!}{
  \begin{tabular}{c|cccc|ccc}
    \toprule
    Authorized/Test & Art & Clipart & Product & RealWorld & $W_{ua} \uparrow$ & $D_u \uparrow$ & $D_a \downarrow$ \\
    \midrule   
        Art & 85.5  $\Rightarrow$  79.5  & 68.0  $\Rightarrow$  52.5  & 89.8  $\Rightarrow$  87.8  & 88.5  $\Rightarrow$  69.8  & 4.82   & 12.07  & 6.00  \\ 
        Clipart & 81.0  $\Rightarrow$  56.0  & 75.0  $\Rightarrow$  75.0  & 90.8  $\Rightarrow$  87.3  & 89.5  $\Rightarrow$  58.5  & 14.88   & 19.83  & 0.00   \\
        Product & 78.8  $\Rightarrow$  46.8  & 73.3  $\Rightarrow$  41.3  & 92.8  $\Rightarrow$  89.0  & 87.5  $\Rightarrow$  60.3  & 23.67   & 30.40  & 3.80  \\ 
        RealWorld & 83.0  $\Rightarrow$  64.8  & 71.3  $\Rightarrow$  35.0  & 90.8  $\Rightarrow$  76.5  & 90.0  $\Rightarrow$  89.8  & 20.41   & 22.93  & 0.20   \\
    \hline
        Mean     & \multicolumn{4}{c|}{/} &15.95 & 21.31 & 2.50 \\  
    \bottomrule
  \end{tabular}}
      \caption{The accuracy ($\%$) of target-free IP-CLIP on the Office-Home-65~[26]. The vertical/horizontal axis denotes the authorized/test domain. In each task, the left of '\(\Rightarrow\)' shows the test accuracy of supervised learning CLIP on the test domain, while the right side presents the accuracy of IP-CLIP. $W_{ua}$ represents the proposed weighted drop, while $D_u$ and $D_a$ denote the drop rates for the unauthorized and authorized domains, respectively.}
  \label{r_tf_home_detail}
\end{table*}

\begin{table*}[!t]
\renewcommand\arraystretch{1.2}
  \centering
  \resizebox{1\textwidth}{!}{
  \begin{tabular}{c|cccc|ccc}
    \toprule
    Authorized/Test & Art & Clipart & Product & Real & $W_{ua} \uparrow$ & $D_u \uparrow$ & $D_a \downarrow$ \\
    \midrule   
        Clipart & 76.4  $\Rightarrow$  59.1  & 49.5  $\Rightarrow$  38.0  & 62.5  $\Rightarrow$  51.6  & 47.0  $\Rightarrow$  34.0  & -3.25   & 11.80  & 17.30  \\
        Painting & 48.0  $\Rightarrow$  36.3  & 61.8  $\Rightarrow$  53.3  & 60.4  $\Rightarrow$  55.3  & 42.3  $\Rightarrow$  36.5  & -0.52   & 7.53  & 8.50  \\
        Real & 52.3  $\Rightarrow$  44.2  & 57.9  $\Rightarrow$  54.1  & 85.6  $\Rightarrow$  83.0  & 46.9  $\Rightarrow$  41.6  & 2.60   & 5.73  & 2.60  \\ 
        Sketch & 52.5  $\Rightarrow$  38.0  & 46.9  $\Rightarrow$  35.5  & 62.8  $\Rightarrow$  45.1  & 66.6  $\Rightarrow$  56.4  & 2.44   & 14.53  & 10.20 \\
    \hline
    Mean      & \multicolumn{4}{c|}{/} & \ 0.32   & 9.90  & 9.47 \\  
    \bottomrule
  \end{tabular}}
      \caption{The accuracy ($\%$) of target-free NTL [27] on the Mini-DomainNet~[32]. The vertical/horizontal axis denotes the authorized/test domain. In each task, the left of '\(\Rightarrow\)' shows the test accuracy of supervised learning CNN on the test domain, while the right side presents the accuracy of NTL. $W_{ua}$ represents the proposed weighted drop, while $D_u$ and $D_a$ denote the drop rates for the unauthorized and authorized domains, respectively.}
  \label{r_tf_mini_ntl}
\end{table*}

\begin{table*}[!t]
\renewcommand\arraystretch{1.2}
  \centering
  \resizebox{1\textwidth}{!}{
  \begin{tabular}{c|cccc|ccc}
    \toprule
    Authorized/Test & Art & Clipart & Product & Real & $W_{ua} \uparrow$ & $D_u \uparrow$ & $D_a \downarrow$ \\
    \midrule   
        Clipart & 76.4  $\Rightarrow$  68.4  & 49.5  $\Rightarrow$  43.9  & 62.5  $\Rightarrow$  57.9  & 47.0  $\Rightarrow$  41.3  & -1.85   & 5.30  & 8.00 \\ 
        Painting & 48.0  $\Rightarrow$  39.3  & 61.8  $\Rightarrow$  58.4  & 60.4  $\Rightarrow$  61.8  & 42.3  $\Rightarrow$  38.0  & 0.27   & 3.87  & 3.40  \\ 
        Real & 52.3  $\Rightarrow$  44.1  & 57.9  $\Rightarrow$  51.2  & 85.6  $\Rightarrow$  82.1  & 46.9  $\Rightarrow$  43.8  & 2.05   & 6.00  & 3.50 \\
        Sketch & 52.5  $\Rightarrow$  44.2  & 46.9  $\Rightarrow$  41.3  & 62.8  $\Rightarrow$  56.6  & 66.7  $\Rightarrow$  57.1  & -1.63   & 6.70  & 9.56  \\ 
    \hline
    Mean      & \multicolumn{4}{c|}{/} & \ -0.29   & 5.47  & 4.97 \\  
    \bottomrule
  \end{tabular}}
      \caption{The accuracy ($\%$) of target-free CUTI-Domain [28] on the Mini-DomainNet~[32]. The vertical/horizontal axis denotes the authorized/test domain. In each task, the left of '\(\Rightarrow\)' shows the test accuracy of supervised learning CNN on the test domain, while the right side presents the accuracy of CUTI-Domain. $W_{ua}$ represents the proposed weighted drop, while $D_u$ and $D_a$ denote the drop rates for the unauthorized and authorized domains, respectively.}
  \label{r_tf_mini_cuti}
\end{table*}

\begin{table*}[!t]
\renewcommand\arraystretch{1.2}
  \centering
  \resizebox{1\textwidth}{!}{
  \begin{tabular}{c|cccc|ccc}
    \toprule
    Authorized/Test & Art & Clipart & Product & Real & $W_{ua} \uparrow$ & $D_u \uparrow$ & $D_a \downarrow$ \\
    \midrule   
        Clipart & 85.1  $\Rightarrow$  80.5  & 79.8  $\Rightarrow$  77.0  & 89.8  $\Rightarrow$  88.4  & 78.7  $\Rightarrow$  72.4  & -0.89   & 3.50  & 4.60 \\
        Painting & 83.8  $\Rightarrow$  78.7  & 81.4  $\Rightarrow$  77.5  & 89.1  $\Rightarrow$  86.4  & 78.4  $\Rightarrow$  73.0  & 0.39   & 4.40  & 3.90 \\ 
        Real & 84.6  $\Rightarrow$  78.9  & 80.5  $\Rightarrow$  69.7  & 90.6  $\Rightarrow$  86.4  & 80.0  $\Rightarrow$  68.4  & 4.46   & 9.37  & 4.20 \\ 
        Sketch & 84.3  $\Rightarrow$  73.2  & 79.1  $\Rightarrow$  74.9  & 90.3  $\Rightarrow$  83.5  & 80.7  $\Rightarrow$  77.3  & 3.07   & 7.37  & 3.40 \\
    \hline
    Mean      & \multicolumn{4}{c|}{/} & \ 1.76   & 6.16  & 4.23 \\  
    \bottomrule
  \end{tabular}}
      \caption{The accuracy ($\%$) of target-free CLIP-based NTL [27] on the Mini-DomainNet~[32]. The vertical/horizontal axis denotes the authorized/test domain. In each task, the left of '\(\Rightarrow\)' shows the test accuracy of supervised learning CLIP on the test domain, while the right side presents the accuracy of CLIP-based NTL. $W_{ua}$ represents the proposed weighted drop, while $D_u$ and $D_a$ denote the drop rates for the unauthorized and authorized domains, respectively.}
  \label{r_tf_mini_ntl_p}
\end{table*}

\begin{table*}[!t]
\renewcommand\arraystretch{1.2}
  \centering
  \resizebox{1\textwidth}{!}{
  \begin{tabular}{c|cccc|ccc}
    \toprule
    Authorized/Test & Art & Clipart & Product & Real & $W_{ua} \uparrow$ & $D_u \uparrow$ & $D_a \downarrow$ \\
    \midrule   
        Clipart & 85.1  $\Rightarrow$  80.8  & 79.8  $\Rightarrow$  67.9  & 89.8  $\Rightarrow$  85.4  & 78.7  $\Rightarrow$  73.8  & 2.24   & 7.07  & 4.30 \\ 
        Painting & 83.8  $\Rightarrow$  80.6  & 81.4  $\Rightarrow$  78.1  & 89.1  $\Rightarrow$  88.9  & 78.4  $\Rightarrow$  71.1  & 0.21   & 3.57  & 3.30 \\ 
        Real & 84.6  $\Rightarrow$  74.3  & 80.5  $\Rightarrow$  74.6  & 90.6  $\Rightarrow$  88.3  & 80.0  $\Rightarrow$  69.4  & 5.86   & 8.93  & 2.30 \\ 
        Sketch & 84.3  $\Rightarrow$  78.9  & 79.1  $\Rightarrow$  74.1  & 90.3  $\Rightarrow$  85.2  & 80.7  $\Rightarrow$  77.2  & 1.29   & 5.17  & 3.50 \\ 
    \hline
    Mean      & \multicolumn{4}{c|}{/} & \ 2.40   & 6.18  & 3.30 \\  
    \bottomrule
  \end{tabular}}
      \caption{The accuracy ($\%$) of target-free CLIP-based CUTI-Domain [28] on the Mini-DomainNet~[32]. The vertical/horizontal axis denotes the authorized/test domain. In each task, the left of '\(\Rightarrow\)' shows the test accuracy of supervised learning CLIP on the test domain, while the right side presents the accuracy of CLIP-based CUTI-Domain. $W_{ua}$ represents the proposed weighted drop, while $D_u$ and $D_a$ denote the drop rates for the unauthorized and authorized domains, respectively.}
  \label{r_tf_mini_cuti_p}
\end{table*}

\begin{table*}[!t]
\renewcommand\arraystretch{1.2}
  \centering
  \resizebox{1\textwidth}{!}{
  \begin{tabular}{c|cccc|ccc}
    \toprule
    Authorized/Test & Art & Clipart & Product & Real & $W_{ua} \uparrow$ & $D_u \uparrow$ & $D_a \downarrow$ \\
    \midrule   
        Clipart & 85.1  $\Rightarrow$  81.1  & 79.8  $\Rightarrow$  70.6  & 89.8  $\Rightarrow$  86.4  & 78.7  $\Rightarrow$  68.4  & 2.95   & 7.63  & 4.00 \\
        Painting & 83.8  $\Rightarrow$  76.5  & 81.4  $\Rightarrow$  78.7  & 89.1  $\Rightarrow$  87.8  & 78.4  $\Rightarrow$  75.2  & 0.97   & 3.93  & 2.70  \\
        Real & 84.6  $\Rightarrow$  66.8  & 80.5  $\Rightarrow$  66.8  & 90.6  $\Rightarrow$  88.1  & 80.0  $\Rightarrow$  57.1  & 13.77   & 18.13  & 2.50 \\
        Sketch & 84.3  $\Rightarrow$  78.6  & 79.1  $\Rightarrow$  69.8  & 90.3  $\Rightarrow$  80.6  & 80.7  $\Rightarrow$  77.3  & 3.74   & 8.23  & 3.40 \\
    \hline
    Mean      & \multicolumn{4}{c|}{/} & \ 5.36   & 9.48  & 3.07 \\  
    \bottomrule
  \end{tabular}}
      \caption{The accuracy ($\%$) of target-free IP-CLIP on the Mini-DomainNet~[32]. The vertical/horizontal axis denotes the authorized/test domain. In each task, the left of '\(\Rightarrow\)' shows the test accuracy of supervised learning CLIP on the test domain, while the right side presents the accuracy of IP-CLIP. $W_{ua}$ represents the proposed weighted drop, while $D_u$ and $D_a$ denote the drop rates for the unauthorized and authorized domains, respectively.}
  \label{r_tf_mini_detail}
\end{table*}

\begin{table*}[!t]
\renewcommand\arraystretch{1.2}
  \centering
  \resizebox{1\textwidth}{!}{
  \begin{tabular}{c|ccc|ccc}
    \toprule
    Authorized/Test & Amazon & Dslr & Webcam & $D_{ua} \uparrow$ & $A_u \downarrow$ & $A_a \uparrow$ \\
    \midrule 
    Amazon & 3.1  & 6.3  & 6.3  & \ 1.63  & 5.21  & 15.63 \\
    Dslr   & 7.8  & 3.1  & 3.1  & \ 9.23  & 4.69  & 32.81 \\
    Webcam & 0.0  & 0.0  & 0.0  &  11.82  & 0.00  & 34.38 \\
    \midrule 
    Mean   & \multicolumn{3}{c|}{/} & 7.56 & 3.30 & 27.60 \\
    \bottomrule
  \end{tabular}}
    \caption{$D_{ua}$, $A_u$, and $A_a$ of authorization application NTL [27] on the Office-31 [25]. The vertical/horizontal axis denotes the authorized/test domain. $D_{ua}$ represents the proposed weighted drop, while $A_u^{IP}$ and $A_u^{IP}$ denote the accuarcy of the unauthorized and test domains, respectively.}
  \label{r_aa_office31_ntl}
\end{table*}

\begin{table*}[!t]
\renewcommand\arraystretch{1.2}
  \centering
  \resizebox{1\textwidth}{!}{
  \begin{tabular}{c|ccc|ccc}
    \toprule
    Authorized/Test & Amazon & Dslr & Webcam & $D_{ua} \uparrow$ & $A_u \downarrow$ & $A_a \uparrow$ \\
    \midrule 
    Amazon & \ 0.0  & \ 1.6 & \ 0.0  & 27.95  & 0.52  & 53.13 \\
    Dslr   &  10.9  & \ 0.0 & \ 1.6  & 72.92  & 4.17  & 87.50 \\
    Webcam &  34.4  &  43.8 &  32.8  & 40.01  & 37.00  & 84.40 \\
    \midrule 
    Mean   & \multicolumn{3}{c|}{/} & 46.96  & 13.90  & 75.01 \\       
    \bottomrule
  \end{tabular}}
    \caption{$D_{ua}$, $A_u$, and $A_a$ of authorization application CUTI-Domain [28] on the Office-31 [25]. The vertical/horizontal axis denotes the authorized/test domain. $D_{ua}$ represents the proposed weighted drop, while $A_u^{IP}$ and $A_u^{IP}$ denote the accuarcy of the unauthorized and test domains, respectively.}
  \label{r_aa_office31_cuti}
\end{table*}

\begin{table*}[!t]
\renewcommand\arraystretch{1.2}
  \centering
  \resizebox{1\textwidth}{!}{
  \begin{tabular}{c|ccc|ccc}
    \toprule
    Authorized/Test & Amazon & Dslr & Webcam & $D_{ua} \uparrow$ & $A_u \downarrow$ & $A_a \uparrow$ \\
    \midrule 
    Amazon & 57.5 & 13.8 & 41.0 & 15.67 & 37.43  & 62.50 \\
    Dslr   & 78.7 & 18.5 & 54.3 & 39.25 & 50.50  & 92.80 \\
    Webcam & 31.8 & 17.3 & 14.8 & 54.59 & 21.30  & 85.30 \\
    \midrule 
    Mean   & \multicolumn{3}{c|}{/} & 36.50 & 36.41 & 80.20 \\
    \bottomrule
  \end{tabular}}
    \caption{$D_{ua}$, $A_u$, and $A_a$ of authorization application CLIP-based NTL [27] on the Office-31 [25]. The vertical/horizontal axis denotes the authorized/test domain. $D_{ua}$ represents the proposed weighted drop, while $A_u^{IP}$ and $A_u^{IP}$ denote the accuarcy of the unauthorized and test domains, respectively.}
  \label{r_aa_office31_ntl_p}
\end{table*}

\begin{table*}[!t]
\renewcommand\arraystretch{1.2}
  \centering
  \resizebox{1\textwidth}{!}{
  \begin{tabular}{c|ccc|ccc}
    \toprule
    Authorized/Test & Amazon & Dslr & Webcam & $D_{ua} \uparrow$ & $A_u \downarrow$ & $A_a \uparrow$ \\
    \midrule 
    Amazon & 36.0  & \ 7.5  & 19.0 & 29.26  & 20.83  & 65.50 \\
    Dslr   & 74.0  & \ 6.3  & 29.3 & 54.47  & 36.53  & 94.30 \\
    Webcam & 57.0  &  12.5  & 22.3 & 40.56  & 30.60  & 80.80 \\
    \midrule 
    Mean   & \multicolumn{3}{c|}{/} & 41.43  & 29.32  & 80.20 \\
    \bottomrule
  \end{tabular}}
    \caption{$D_{ua}$, $A_u$, and $A_a$ of authorization application CLIP-based CUTI-Domain [28] on the Office-31 [25]. The vertical/horizontal axis denotes the authorized/test domain. $D_{ua}$ represents the proposed weighted drop, while $A_u^{IP}$ and $A_u^{IP}$ denote the accuarcy of the unauthorized and test domains, respectively.}
  \label{r_aa_office31_cuti_p}
\end{table*}

\begin{table*}[!t]
\renewcommand\arraystretch{1.2}
  \centering
  \resizebox{1\textwidth}{!}{
  \begin{tabular}{c|ccc|ccc}
    \toprule
    Authorized/Test & Amazon & Dslr & Webcam & $D_{ua} \uparrow$ & $A_u \downarrow$ & $A_a \uparrow$ \\
    \midrule 
    Amazon & \ 4.5 & 3.3 & \ 2.8 & 37.46 & \ 3.53 & 63.00 \\
    Dslr   &  27.3 & 1.5 & \ 0.5 & 82.42 & \ 9.77 & 95.80 \\
    Webcam &  31.0 & 4.3 &  11.3 & 56.45 &  15.53 & 83.30 \\
    \midrule 
    Mean   & \multicolumn{3}{c|}{/} & 58.78 & \ 9.61 & 80.70 \\
    \bottomrule
  \end{tabular}}
    \caption{$D_{ua}$, $A_u$, and $A_a$ of authorization application IP-CLIP on the Office-31 [25]. The vertical/horizontal axis denotes the authorized/test domain. $D_{ua}$ represents the proposed weighted drop, while $A_u^{IP}$ and $A_u^{IP}$ denote the accuarcy of the unauthorized and test domains, respectively.}
  \label{r_aa_office31_detail}
\end{table*}

\begin{table*}[!t]
\renewcommand\arraystretch{1.2}
  \centering
  \resizebox{1\textwidth}{!}{
  \begin{tabular}{c|cccc|ccc}
    \toprule
    Authorized/Test & Art & Clipart & Product & RealWorld & $D_{ua} \uparrow$ & $A_u \downarrow$ & $A_a \uparrow$ \\
    \midrule 
    Art       & 77.3  & 42.7  & 64.6  & 71.1  & 8.75  & 63.93  & 75.52 \\
    Clipart   & 44.8  & 58.9  & 46.4  & 51.6  & 4.98  & 50.39  & 58.85 \\
    Product   & 50.0  & 43.0  & 78.4  & 62.2  & 17.49  & 58.40  & 80.21 \\
    RealWorld & 60.9  & 46.6  & 69.3  & 83.1  & 15.83  & 64.97  & 83.85 \\
    \midrule 
    Mean      & \multicolumn{4}{c|}{/} & 11.76  & 59.42  & 74.61 \\
    \bottomrule
  \end{tabular}}
      \caption{$D_{ua}$, $A_u$, and $A_a$ of authorization application NTL [27] on the Office-Home-65~[26]. The vertical/horizontal axis denotes the authorized/test domain. $D_{ua}$ represents the proposed weighted drop, while $A_u^{IP}$ and $A_u^{IP}$ denote the accuarcy of the unauthorized and test domains, respectively.}
  \label{r_aa_home_ntl}
\end{table*}

\begin{table*}[!t]
\renewcommand\arraystretch{1.2}
  \centering
  \resizebox{1\textwidth}{!}{
  \begin{tabular}{c|cccc|ccc}
    \toprule
    Authorized/Test & Art & Clipart & Product & RealWorld & $D_{ua} \uparrow$ & $A_u \downarrow$ & $A_a \uparrow$ \\
    \midrule 
    Art       & \ 1.6  & \ 1.0  & \ 0.8  & \ 0.8  &  35.25  & \ 1.04 & 59.90 \\
    Clipart   & \ 0.8  & \ 0.8  & \ 0.8  & \ 0.5  &  14.78  & \ 0.72 & 38.80 \\
    Product   & \ 2.1  & \ 0.8  & \ 0.0  & \ 0.3  &  33.27  & \ 0.78 & 58.07 \\
    RealWorld &  21.9  &  19.5  &  39.6  &  44.3  & \ 3.15  &  31.32 & 39.32 \\
    \midrule 
    Mean      & \multicolumn{4}{c|}{/} & 21.61 & 8.46 & 49.02 \\
    \bottomrule
  \end{tabular}}
      \caption{$D_{ua}$, $A_u$, and $A_a$ of authorization application CUTI-Domain [28] on the Office-Home-65~[26]. The vertical/horizontal axis denotes the authorized/test domain. $D_{ua}$ represents the proposed weighted drop, while $A_u^{IP}$ and $A_u^{IP}$ denote the accuarcy of the unauthorized and test domains, respectively.}
  \label{r_aa_home_cuti}
\end{table*}

\begin{table*}[!t]
\renewcommand\arraystretch{1.2}
  \centering
  \resizebox{1\textwidth}{!}{
  \begin{tabular}{c|cccc|ccc}
    \toprule
    Authorized/Test & Art & Clipart & Product & RealWorld & $D_{ua} \uparrow$ & $A_u \downarrow$ & $A_a \uparrow$ \\
    \midrule 
    Art       & 21.0 & 14.3 & 17.0 & 29.5 &  49.47 & 20.45 & 81.30 \\
    Clipart   & 21.0 & 13.0 & 45.0 & 31.8 & \ 9.74 & 27.70 & 48.00 \\
    Product   & 21.3 & 27.8 & 36.8 & 24.0 &  44.44 & 27.48 & 81.80 \\
    RealWorld & 10.3 & 22.5 & 27.5 & 16.5 &  51.50 & 19.20 & 82.00 \\
    \midrule 
    Mean      & \multicolumn{4}{c|}{/} & 38.79 & 23.71 & 73.28 \\
    \bottomrule
  \end{tabular}}
      \caption{$D_{ua}$, $A_u$, and $A_a$ of authorization application CLIP-based NTL [27] on the Office-Home-65~[26]. The vertical/horizontal axis denotes the authorized/test domain. $D_{ua}$ represents the proposed weighted drop, while $A_u^{IP}$ and $A_u^{IP}$ denote the accuarcy of the unauthorized and test domains, respectively.}
  \label{r_aa_home_ntl_p}
\end{table*}

\begin{table*}[!t]
\renewcommand\arraystretch{1.2}
  \centering
  \resizebox{1\textwidth}{!}{
  \begin{tabular}{c|cccc|ccc}
    \toprule
    Authorized/Test & Art & Clipart & Product & RealWorld & $D_{ua} \uparrow$ & $A_u \downarrow$ & $A_a \uparrow$ \\
    \midrule 
    Art       & \ 4.5  & \ 5.0 & 21.0 &  11.0 & 54.95 &  10.38 & 79.50 \\
    Clipart   &  11.0  &  16.0 & 36.5 &  20.0 & 16.86 &  20.88 & 52.80 \\
    Product   &  18.0  &  33.5 & 61.0 &  29.0 & 39.53 &  35.38 & 83.00 \\
    RealWorld & \ 7.5  & \ 3.5 & 10.8 & \ 9.5 & 62.87 & \ 7.83 & 83.30 \\
    \midrule 
    Mean      & \multicolumn{4}{c|}{/} & 43.55 & 18.61 & 74.65 \\
    \bottomrule
  \end{tabular}}
      \caption{$D_{ua}$, $A_u$, and $A_a$ of authorization application CLIP-based CUTI-Domain [28] on the Office-Home-65~[26]. The vertical/horizontal axis denotes the authorized/test domain. $D_{ua}$ represents the proposed weighted drop, while $A_u^{IP}$ and $A_u^{IP}$ denote the accuarcy of the unauthorized and test domains, respectively.}
  \label{r_aa_home_cuti_p}
\end{table*}

\begin{table*}[!t]
\renewcommand\arraystretch{1.2}
  \centering
  \resizebox{1\textwidth}{!}{
  \begin{tabular}{c|cccc|ccc}
    \toprule
    Authorized/Test & Art & Clipart & Product & RealWorld & $D_{ua} \uparrow$ & $A_u \downarrow$ & $A_a \uparrow$ \\
    \midrule 
    Art       & 1.5 & 3.3 & \ 7.8 & 3.0 & 60.12 & \ 3.88 & 79.50 \\
    Clipart   & 4.3 & 5.3 &  22.5 & 9.8 & 26.52 &  10.48 & 57.00 \\
    Product   & 5.8 & 9.3 &  12.0 & 6.5 & 57.74 & \ 8.40 & 80.30 \\
    RealWorld & 2.3 & 4.0 & \ 8.5 & 6.0 & 71.17 & \ 5.20 & 87.00 \\
    \midrule 
    Mean      & \multicolumn{4}{c|}{/} & 53.89 & \ 6.99 & 75.95 \\
    \bottomrule
  \end{tabular}}
      \caption{$D_{ua}$, $A_u$, and $A_a$ of authorization application IP-CLIP on the Office-Home-65~[26]. The vertical/horizontal axis denotes the authorized/test domain. $D_{ua}$ represents the proposed weighted drop, while $A_u^{IP}$ and $A_u^{IP}$ denote the accuarcy of the unauthorized and test domains, respectively.}
  \label{r_aa_home_detail}
\end{table*}

\begin{table*}[!t]
\renewcommand\arraystretch{1.2}
  \centering
  \resizebox{1\textwidth}{!}{
  \begin{tabular}{c|cccc|ccc}
    \toprule
    Authorized/Test & Clipart & Painting & Real & Sketch & $D_{ua} \uparrow$ & $A_u \downarrow$ & $A_a \uparrow$ \\
    \midrule 
    Clipart   & 75.0 & 48.2 & 62.2 & 46.9 &  11.96 & 58.06 & 74.18 \\
    Painting  & 51.6 & 69.1 & 68.4 & 43.9 & \ 7.47 & 58.26 & 69.08 \\
    Real      & 47.5 & 53.1 & 83.2 & 44.2 &  21.08 & 57.03 & 82.57 \\
    Sketch    & 53.9 & 48.4 & 62.7 & 68.9 & \ 7.72 & 58.47 & 69.57 \\
    \midrule 
    Mean      & \multicolumn{4}{c|}{/} & 12.06 & 57.96 & 73.85 \\
    \bottomrule
  \end{tabular}}
      \caption{$D_{ua}$, $A_u$, and $A_a$ of authorization application NTL [27] on the Mini-DomainNet [31]. The vertical/horizontal axis denotes the authorized/test domain. $D_{ua}$ represents the proposed weighted drop, while $A_u^{IP}$ and $A_u^{IP}$ denote the accuarcy of the unauthorized and test domains, respectively.}
  \label{r_aa_mini_ntl}
\end{table*}

\begin{table*}[!t]
\renewcommand\arraystretch{1.2}
  \centering
  \resizebox{1\textwidth}{!}{
  \begin{tabular}{c|cccc|ccc}
    \toprule
    Authorized/Test & Clipart & Painting & Real & Sketch & $D_{ua} \uparrow$ & $A_u \downarrow$ & $A_a \uparrow$ \\
    \midrule 
    Clipart   & 78.5 & 50.5 & 64.1 & 49.0 &  13.75 & 60.53 & 78.13 \\
    Painting  & 38.5 & 56.3 & 54.6 & 31.3 & \ 6.47 & 45.15 & 56.58 \\
    Real      & 48.5 & 54.9 & 85.4 & 44.9 &  22.62 & 58.43 & 85.03 \\
    Sketch    & 50.8 & 48.8 & 61.0 & 68.3 & \ 7.00 & 57.24 & 67.60 \\
    \midrule 
    Mean      & \multicolumn{4}{c|}{/} & 12.46 & 55.34 & 71.83 \\
    \bottomrule        
  \end{tabular}}
      \caption{$D_{ua}$, $A_u$, and $A_a$ of authorization application CUTI-Domain [28] on the Mini-DomainNet [31]. The vertical/horizontal axis denotes the authorized/test domain. $D_{ua}$ represents the proposed weighted drop, while $A_u^{IP}$ and $A_u^{IP}$ denote the accuarcy of the unauthorized and test domains, respectively.}
  \label{r_aa_mini_cuti}
\end{table*}

\begin{table*}[!t]
\renewcommand\arraystretch{1.2}
  \centering
  \resizebox{1\textwidth}{!}{
  \begin{tabular}{c|cccc|ccc}
    \toprule
    Authorized/Test & Clipart & Painting & Real & Sketch & $D_{ua} \uparrow$ & $A_u \downarrow$ & $A_a \uparrow$ \\
    \midrule 
    Clipart   & 13.5 & \ 8.7 & 12.1 & 35.9 & 38.45 & 17.54 & 71.40 \\
    Painting  & 26.2 &  15.1 & 14.3 & 41.1 & 32.78 & 24.18 & 70.60 \\
    Real      & 41.1 &  21.8 & 24.6 & 64.1 & 35.66 & 37.90 & 81.60 \\
    Sketch    & 16.5 & \ 7.8 & 11.3 & 30.6 & 38.66 & 16.55 & 71.00 \\
    \midrule 
    Mean      & \multicolumn{4}{c|}{/} & 36.39 & 24.04 & 73.65 \\       
    \bottomrule
  \end{tabular}}
      \caption{$D_{ua}$, $A_u$, and $A_a$ of authorization application CLIP-based NTL [27] on the Mini-DomainNet [31]. The vertical/horizontal axis denotes the authorized/test domain. $D_{ua}$ represents the proposed weighted drop, while $A_u^{IP}$ and $A_u^{IP}$ denote the accuarcy of the unauthorized and test domains, respectively.}
  \label{r_aa_mini_ntl_p}
\end{table*}

\begin{table*}[!t]
\renewcommand\arraystretch{1.2}
  \centering
  \resizebox{1\textwidth}{!}{
  \begin{tabular}{c|cccc|ccc}
    \toprule
    Authorized/Test & Clipart & Painting & Real & Sketch & $D_{ua} \uparrow$ & $A_u \downarrow$ & $A_a \uparrow$ \\
    \midrule 
    Clipart   &  57.9 &  24.3 &  31.1 & 63.0 & 22.77 &  44.08 & 74.60 \\
    Painting  &  46.5 &  13.8 &  21.6 & 57.0 & 24.48 &  34.73 & 69.80 \\
    Real      &  42.4 &  17.9 &  21.9 & 57.1 & 33.56 &  34.83 & 77.90 \\
    Sketch    & \ 6.0 & \ 4.8 & \ 6.4 & 20.6 & 48.18 & \ 9.45 & 74.30 \\
    \midrule 
    Mean      & \multicolumn{4}{c|}{/} & 32.25 & 30.77 & 74.15 \\
    \bottomrule
  \end{tabular}}
      \caption{$D_{ua}$, $A_u$, and $A_a$ of authorization application CLIP-based CUTI-Domain [28] on the Mini-DomainNet [31]. The vertical/horizontal axis denotes the authorized/test domain. $D_{ua}$ represents the proposed weighted drop, while $A_u^{IP}$ and $A_u^{IP}$ denote the accuarcy of the unauthorized and test domains, respectively.}
  \label{r_aa_mini_cuti_p}
\end{table*}

\begin{table*}[!t]
\renewcommand\arraystretch{1.2}
  \centering
  \resizebox{1\textwidth}{!}{
  \begin{tabular}{c|cccc|ccc}
    \toprule
    Authorized/Test & Clipart & Painting & Real & Sketch & $D_{ua} \uparrow$ & $A_u \downarrow$ & $A_a \uparrow$ \\
    \midrule 
    Clipart   & \ 4.6 & 5.1 & 4.8 & 14.9 & 50.88 & \ 7.35 & 75.10 \\
    Painting  & \ 7.5 & 4.6 & 8.4 & 23.2 & 40.33 &  10.93 & 69.20 \\
    Real      &  17.3 & 8.4 & 9.8 & 38.1 & 54.06 &  18.40 & 83.30 \\
    Sketch    & \ 4.6 & 3.8 & 5.7 & 18.7 & 48.27 & \ 8.20 & 73.70 \\
    \midrule 
    Mean      & \multicolumn{4}{c|}{/} & 48.39 &  11.22 & 75.33 \\
    \bottomrule
  \end{tabular}}
      \caption{$D_{ua}$, $A_u$, and $A_a$ of authorization application IP-CLIP on the Mini-DomainNet [31]. The vertical/horizontal axis denotes the authorized/test domain. $D_{ua}$ represents the proposed weighted drop, while $A_u^{IP}$ and $A_u^{IP}$ denote the accuarcy of the unauthorized and test domains, respectively.}
  \label{r_aa_mini_detail}
\end{table*}

\begin{table*}[!t]
\renewcommand\arraystretch{1.2}
  \centering
  \resizebox{1\textwidth}{!}{
  \begin{tabular}{c|cccc|ccc}
    \toprule
    Modules &  \(\mathcal{L}_a\) & \(\mathcal{L}_u\) & \(\mathcal{L}_{dis}\) & \(\mathcal{L}_{aug}\) & \(W_{ua}\uparrow\) & \(D_u\uparrow\) & \(D_a\downarrow\)  \\
    \midrule
    Baseline (SL-CLIP) & \checkmark & & & & / & / & / \\
    Baseline+IP & \checkmark & \checkmark & & & 23.96 & 30.81 & 2.35 \\
    Baseline+IP+Proj & \checkmark & \checkmark & \checkmark & & 53.60 & 65.41 & 0.41 \\
    \textbf{Proposed} (Baseline+IP+Proj+STAM) & \checkmark & \checkmark & \checkmark & \checkmark & \textbf{54.68} & \textbf{65.48} & \textbf{0.33} \\
    \bottomrule
  \end{tabular}}
      \caption{Ablation experiments on Mini-DomainNet. \(\mathcal{L}_a\) is used for supervised learning on the authorized domain as the baseline. "Baseline+IP" naively trains both domains simultaneously, using entropy ($\mathcal{L}_{en}$) to enhance text feature diversity. The addition of the IP projector (Baseline+IP+Proj) incorporates $\mathcal{L}_{dis}~(\mathcal{L}_{kl} + \mathcal{L}_m)$ to distinguishing text and domain features across domains. STAM with $\mathcal{L}_{aug}~(\mathcal{L}_{ai} + \mathcal{L}_{ui})$ further enhances domain token robustness in domain feature identification (Proposed).}
  \label{r_ablation_mini_detail}
\end{table*}